%% file: neurips_2026_v3.tex
\documentclass{article}

\PassOptionsToPackage{numbers, compress}{natbib}
\usepackage[main, final]{neurips_2026}
\makeatletter
\renewcommand{\@noticestring}{Accepted to the 40th Conference on Neural Information Processing Systems (NeurIPS 2026).}
\makeatother

\usepackage[utf8]{inputenc} 
\usepackage[T1]{fontenc}    
\usepackage{hyperref}       
\usepackage{url}            
\usepackage{booktabs}       
\usepackage{amsfonts}       
\usepackage{nicefrac}       
\usepackage{microtype}      
\usepackage{xcolor}         
\usepackage{graphicx}
\usepackage{subcaption}
\usepackage{amsmath}
\usepackage{multirow}
\usepackage{booktabs}
\usepackage{adjustbox}
\usepackage{makecell}
\usepackage{enumitem}
\usepackage{placeins}
\usepackage{algorithm}
\usepackage{algpseudocode}
\usepackage{fontawesome5}
\hypersetup{colorlinks=true, urlcolor=blue}

\algnewcommand\Input{\item[\textbf{Input:}]}
\algnewcommand\Output{\item[\textbf{Output:}]}
\title{ProCTI: Prototype-Refined Global Conditioning for Diffusion-Based Time Series Imputation}

\author{%
  Fariza Rashid\textsuperscript{1}\thanks{\raggedright Corresponding authors: \texttt{fariza.rashid@sydney.edu.au, suranga.seneviratne@sydney.edu.au, g.batista@unsw.edu.au}} \quad
  Duc Van Le\textsuperscript{2} \quad
  Rahat Masood\textsuperscript{2} \quad \\ [0.5em]
  \textbf{Gustavo Batista}\textsuperscript{2}  \quad
  \textbf{Aruna Seneviratne}\textsuperscript{2} \quad
  \textbf{Suranga Seneviratne}\textsuperscript{1} \\[0.5em]
  \textsuperscript{1}University of Sydney \quad
  \textsuperscript{2}University of New South Wales
}

\begin{document}

\maketitle

\begin{abstract}
Time series imputation has progressed from statistical and deep learning approaches to diffusion-based models, which have shown strong recent performance. Existing diffusion-based methods typically condition the reverse process using local contextual information from the current or neighbouring windows. Meanwhile, global dataset-level structure often remains implicit, limiting performance when local observations are sparse, noisy, or unrepresentative. To address this issue, we propose ProCTI, a diffusion-imputation framework that augments local conditioning with retrieved global dataset-level priors through learned prototypes. A hybrid conditioning mechanism integrates this global context with local signals during reverse diffusion, enabling more accurate reconstruction under varying missingness scenarios. Experiments across multiple benchmark datasets show that ProCTI outperforms strong baselines overall under random missingness, while remaining competitive under attribute-wise missingness. Furthermore, we use a latent-regime data model to characterise the precise conditions under which prototype-derived global conditioning provably improves imputation. We support this with a general theoretical analysis of local-global conditioning. 

\vspace{0.5em}
\begin{center}
\faGithub\ \href{https://github.com/fariza25/ProCTI}{ProCTI Code Repository}
\end{center}

\end{abstract}

\input{1_introduction_v3}
\input{2_relatedwork}

\input{3_framework_v3}

\input{4_experiments_v3}
\input{5_conclusion_v3}

\begin{ack}
This research is funded by the Office of National Intelligence, Australia (Agreement No.\ GA396359), and the National Intelligence and Security Discovery Research Grant (NI240100159).
\end{ack}

\newpage

\bibliographystyle{unsrtnat}
\bibliography{references}

\newpage
\input{appendix_v3}

\FloatBarrier
\clearpage


\end{document}

%% file: 1_introduction_v3.tex
\section{Introduction}

Time series data appears in a wide range of real-world applications, including weather forecasting, finance, healthcare, cybersecurity, human-activity recognition, and audio signal processing~\cite{jena_climate, google_stock_yahoo, reyna2020early, ngo2014largest, gunawardena2024single}. Incomplete or missing data due to sensor malfunctions, human error, or communication failures is therefore a significant concern as it can degrade predictive performance and lead to harmful downstream decisions~\cite{alwan2022time, du2020missing}. Besides statistical, machine learning and deep learning methods~\cite{amiri2016missing, van2011mice, liu2023multivariate, cao2018brits, nie2024imputeformer, fortuin2020gp}, diffusion probabilistic models have emerged as a powerful paradigm for time series imputation~\cite{yuan2024diffusion, tashiro2021csdi}. While existing diffusion-based approaches typically condition the reverse denoising process using signals derived from local context~\cite{tashiro2021csdi, zhou2024mtsci, yang2024frequency}, broader dataset-level patterns are only implicitly encoded in learned parameters. Consequently, when local observations are sparse or noisy, the available context may be insufficient to accurately reflect the true underlying distribution, increasing uncertainty in the denoising process. Diffusion models conditioned only on local context may therefore struggle to generate accurate reconstructions of missing data.

In light of this limitation, we propose ProCTI, a diffusion-based time series imputation framework that explicitly incorporates dataset-level signals into the denoising process. Specifically, we aim to capture recurring global patterns of multivariate temporal behaviour that may exist across the dataset. We refer to these patterns as regimes. To approximate such regimes, ProCTI introduces a learnable prototype bank within a prototype-conditioned diffusion model. Each partially observed input window queries the prototype bank to retrieve a window-specific global context vector, which refines the conditioning signal throughout the reverse diffusion process. This mechanism enables the model to repeatedly incorporate global structural priors during denoising, producing more accurate and coherent imputations.

\begin{figure*}[t]
\centering

\newcommand{\wa}{0.31}
\newcommand{\wb}{0.31}
\newcommand{\wc}{0.367}

\begin{minipage}[t]{\wa\textwidth}
    \centering
    \includegraphics[width=\linewidth]{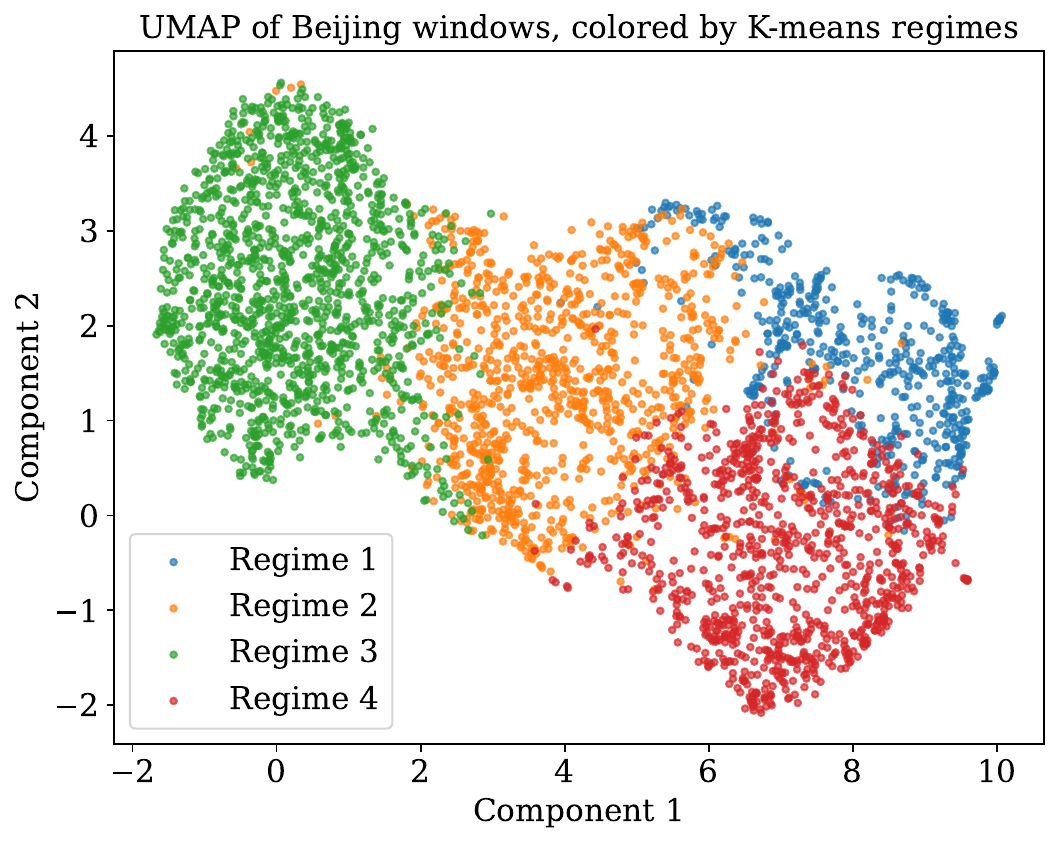}
    \subcaption{}
\end{minipage}
\hfill
\begin{minipage}[t]{\wb\textwidth}
    \centering
    \includegraphics[width=\linewidth]{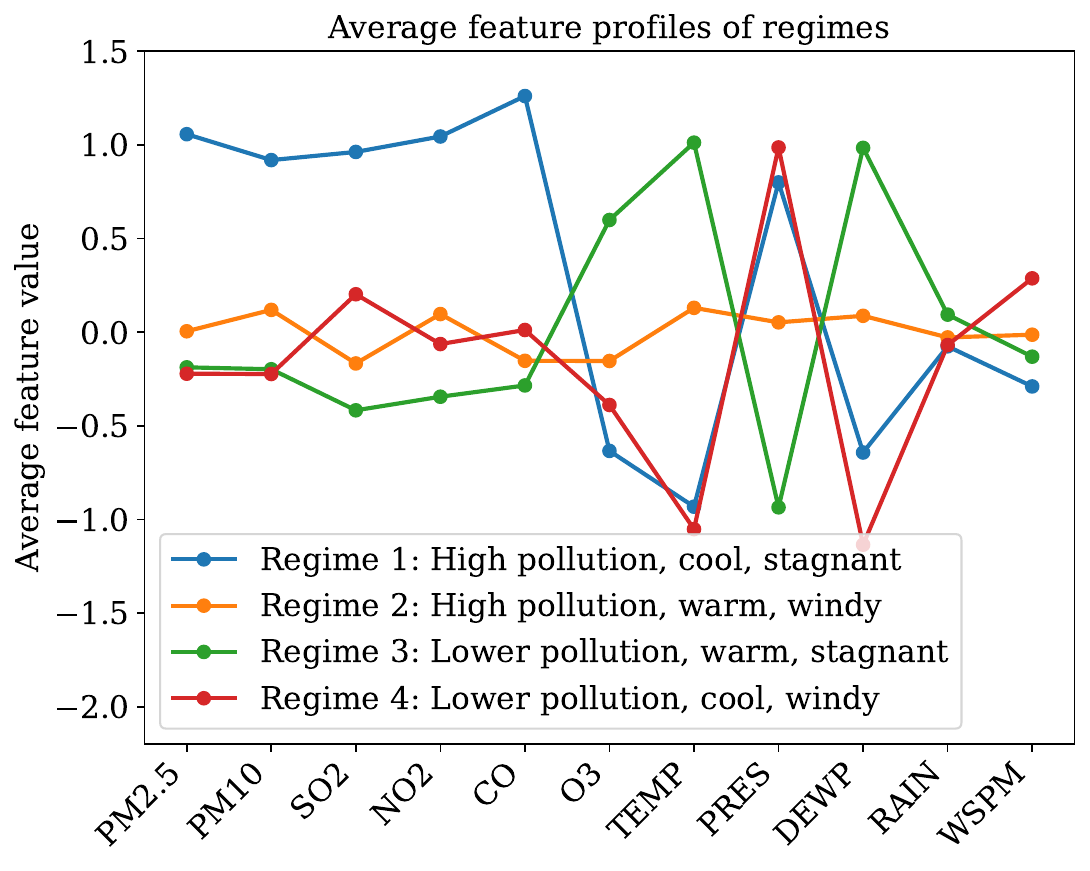}
    \subcaption{}
\end{minipage}
\hfill
\begin{minipage}[t]{\wc\textwidth}
    \centering
    \includegraphics[width=\linewidth]{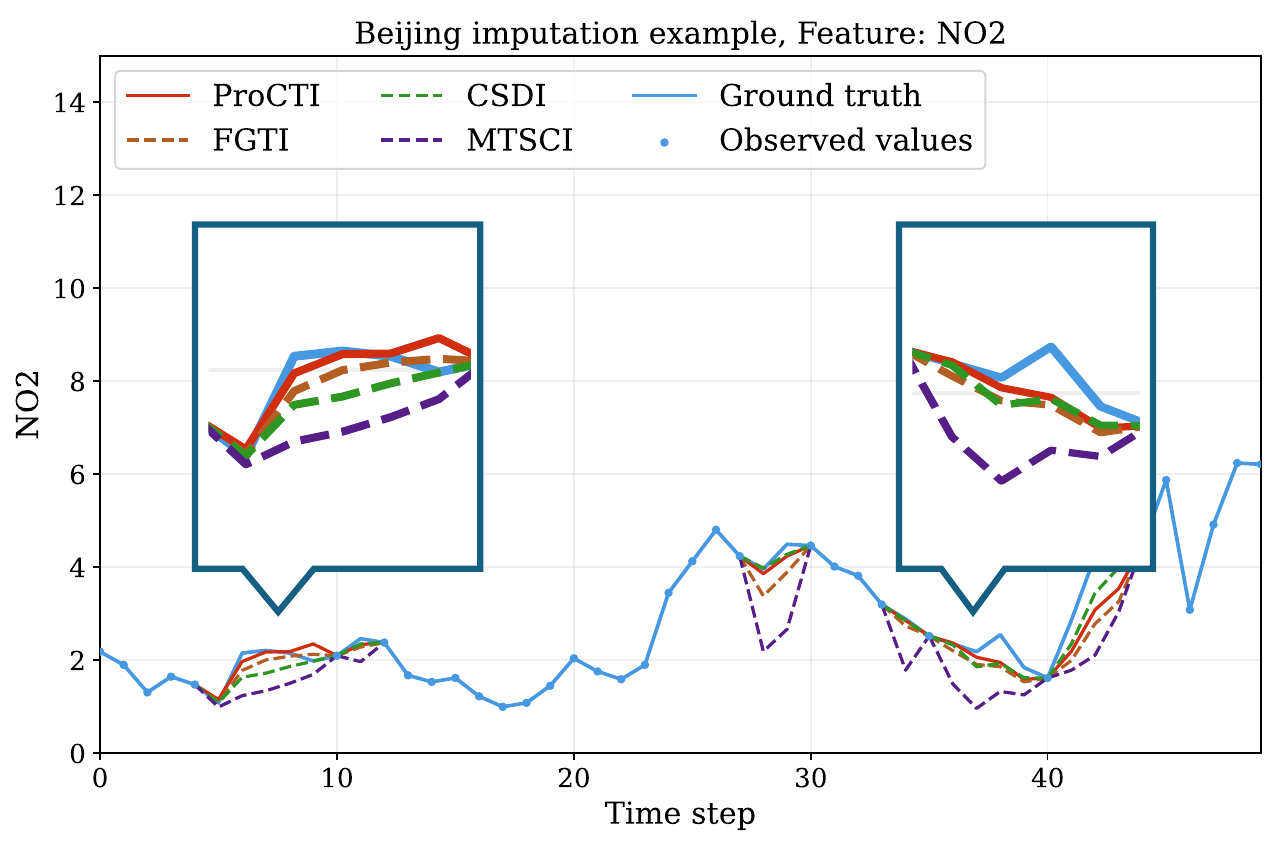}
    \subcaption{}
\end{minipage}

\caption{
Regime structure and representative imputation behaviour in the Beijing dataset.
(a) UMAP projection of normalised windows after PCA, coloured by K-means derived regimes.
(b) Mean feature profiles of the discovered regimes, showing distinct combinations of pollution and meteorological conditions.
(c) Example masked-window reconstruction for NO2, where ProCTI better matches the ground truth than competing baselines.
}
\label{fig:intro_regimes}
\end{figure*}

As a pre-hoc motivating analysis, in Figure~\ref{fig:intro_regimes} we present a representative illustration of global regimes using the Beijing Air Quality dataset~\cite{beijing_multi-site_air_quality_501}. The left panel presents a UMAP visualisation of z-score normalised fixed-length windows, where each window is flattened and projected into a lower-dimensional space using PCA. Applying K-means clustering in this PCA-reduced space reveals four coarse groups of windows. The middle panel shows the corresponding average feature profiles, obtained by averaging each feature across all windows assigned to a given cluster. Because these profiles are derived from samples spanning the full dataset, they reveal recurring regimes characterised by different combinations of pollution indicators (PM2.5, PM10, SO2, NO2), temperature (TEMP), and wind speed (WSPM). Such dataset-level structure may be difficult to infer from a partially observed window alone. The right panel presents a masked-window imputation example, where leveraging this additional global context enables ProCTI to achieve lower MAE than competing baselines.

Overall, we make the following contributions:
\begin{itemize}[noitemsep, topsep=0pt]
    \item We introduce ProCTI, a new conditioning framework for diffusion-based time series imputation that explicitly combines retrieved global dataset-level priors with local context. We implement this framework through a prototype-conditioned diffusion model that improves imputation accuracy.

    \item We present a theoretical perspective that distinguishes local and global conditioning, using a latent-regime data model to characterise the precise conditions under which prototype-derived global conditioning provably improves imputation. This builds on standard supporting theoretical results that connect global conditioning to the diffusion training objective.

    \item We evaluate ProCTI against nine baselines on five real-world datasets: ProCTI achieves the best mean MAE on all five datasets under random missingness, reducing MAE by approximately $3$--$30\%$ over the strongest diffusion baseline (FGTI), and is best or second-best in most attribute-wise missingness scenarios.

\end{itemize}

%% file: 2_relatedwork.tex
\section{Related Work}

\textbf{Time series modelling} Time-series modelling has seen rapid progress across generation, forecasting, and imputation tasks, with early approaches relying on statistical techniques and autoregressive methods~\cite{bartholomew1971time,watson1994vector,amiri2016missing}. More recent machine learning and deep learning methods emerged such as TIDER~\cite{liu2023multivariate}, which uses matrix factorisation to disentangle temporal components, and BRITS~\cite{cao2018brits} and SAITS~\cite{du2023saits}, which employ bidirectional Recurrent Neural Networks (RNNs) and self-attention respectively, to capture long-range dependencies. In forecasting, non-Transformer approaches such as SCINet~\cite{liu2022scinet} and FreTS~\cite{yi2023frequency} model temporal interactions through recursive convolution and frequency-domain multilayer perceptrons respectively. Transformer-based forecasting methods~\cite{zhou2021informer, wu2021autoformer, liu2021pyraformer, liu2022non, nie2022time, liu2023itransformer} have also demonstrated strong performance by modeling long-range temporal dependencies through attention-based architectures, while time series generation methods like Diffusion-TS~\cite{yuan2024diffusion} and PaD-TS~\cite{li2025population} are diffusion-based models which generate realistic synthetic sequences by learning temporal dynamics and preserving dataset-level statistical properties.

\textbf{Conditioning in imputation models} Existing diffusion-based imputation methods incorporate contextual information in varied ways, mostly relying on local or task-specific priors rather than explicitly learning dataset-level global structure. CSDI~\cite{tashiro2021csdi} conditions the reverse process directly on observed values and attention-based temporal/feature dependencies, without an explicit global prior. PriSTI~\cite{liu2023pristi} introduces context for spatiotemporal data by extracting coarse global priors from observed values together with geographic relationships, but this prior is specific to spatial sensor networks rather than general multivariate time series. MTSCI~\cite{zhou2024mtsci} conditions on neighbouring-window information and consistency constraints, thereby exploiting short-range temporal context instead of dataset-wide structure, while FGTI~\cite{yang2024frequency} incorporates frequency-domain priors through dominant- and high-frequency signals. Non-diffusion models such as TG-MSFM~\cite{wangtime} use multi-scale flow trajectories to reconcile global trends with local details during deterministic reconstruction.

In contrast, we formalise conditioning for diffusion-based imputation by distinguishing \emph{local context}, derived from the partially observed input window, and \emph{global context}, representing recurring dataset-level structure shared across samples. We define their combination within the reverse diffusion process as complementary sources of guidance for reconstruction. Following this principle, ProCTI leverages a learned prototype bank to retrieve and inject global context alongside local observations into the denoising process for conditional imputation. This differs from PaD-TS~\cite{li2025population}, which preserves population-level statistics for unconditional time-series generation. We provide an overview of existing prototype-based methods in time-series modelling in Appendix~\ref{appn:relatedwork}.

%% file: 3_framework_v3.tex
\section{Prototype-Refined Global Conditioning}

\begin{figure*}[t]
    \centering
    \includegraphics[width=\textwidth]{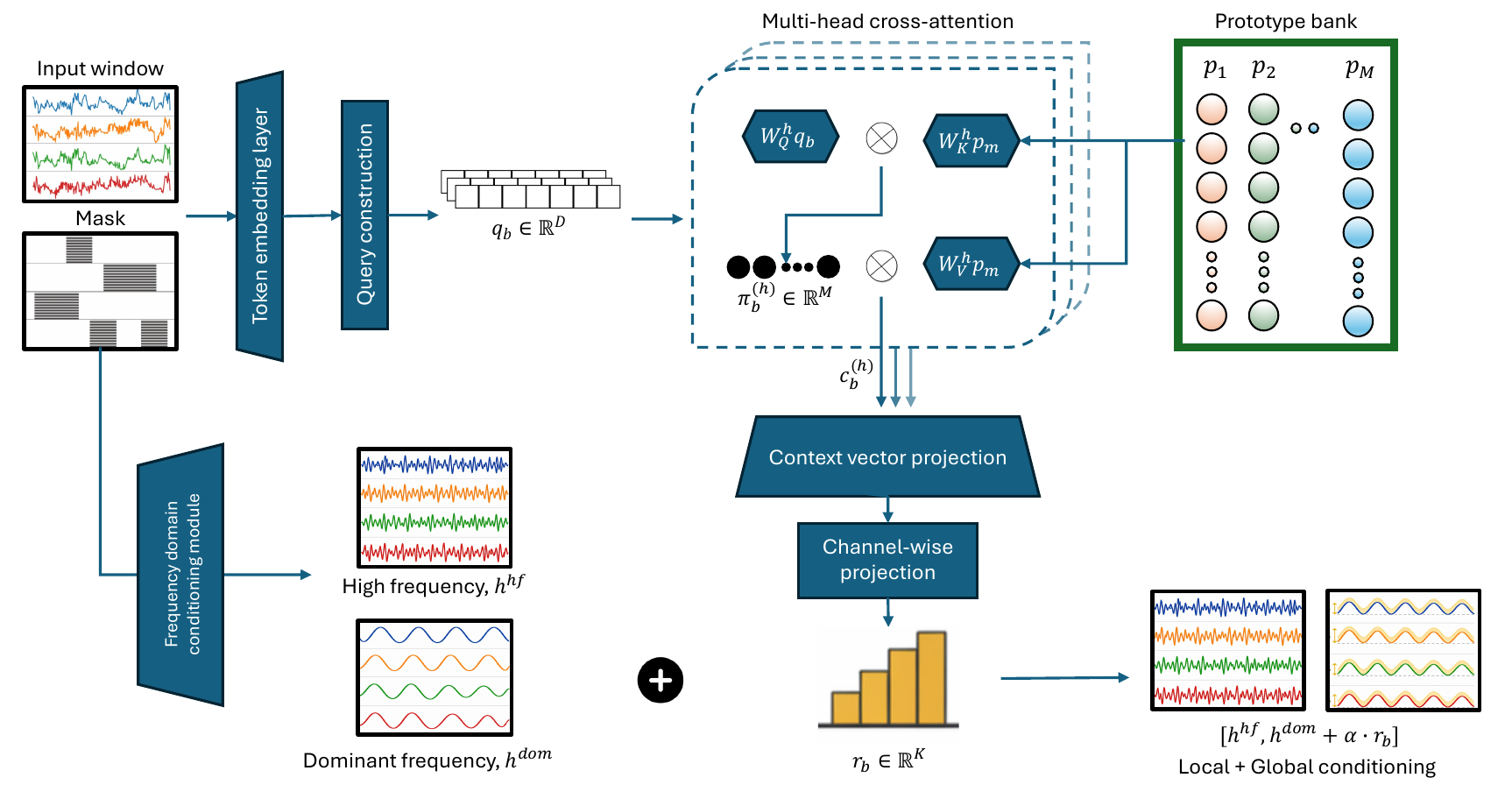}
    \caption{Prototype-conditioning module}
    \label{fig:proto_conditioning_module}
\end{figure*}

In this section, we present ProCTI, a diffusion-imputation framework that combines local window evidence with retrieved global dataset-level priors during reverse diffusion. As illustrated in Figure~\ref{fig:proto_conditioning_module}, ProCTI learns dataset-level prototypes that approximate recurring global regimes and uses them to produce instance-specific (i.e., window-level) conditioning signals. We first describe the prototype conditioning module, which combines global prototype guidance with local contextual information. We then formally distinguish local and global conditioning and use a latent-regime data model to characterise when combining them provably improves imputation. This builds on standard supporting results connecting global conditioning to the diffusion training objective.

\subsection{Prototype-conditioning module}\label{proto_module}

The prototype conditioning module contains a learnable prototype bank trained over the training dataset to approximate recurring global regimes. For each input window, cross-attention retrieves an attribute-wise global context vector from the prototype bank. This signal is then combined with local contextual information to form the conditioning input for the reverse diffusion process.

\noindent\textbf{Problem setup.}
We consider a multivariate time series dataset $X$ consisting of windows $x \in \mathbb{R}^{L\times K}$, where $L$ represents the time length of the window and $K$ represents the number of attributes. While the window may have naturally missing values, we apply an artificial keep-mask $m \in \{0,1\}^{L \times K}$, where $m_{t,k}=1$ indicates that the value at time step $t$ and attribute $k$ is observed, and $m_{t,k}=0$ otherwise. During training, the model learns to reconstruct values at masked positions through a diffusion denoising process.

\paragraph{Token construction.}
Given a mini-batch of $B$ input windows $\mathbf{X} \in \mathbb{R}^{B \times L \times K}$ and the corresponding keep-masks $\mathbf{M} \in \{0,1\}^{B \times L \times K}$, where $B$ denotes the batch size, we construct a 2-dimensional token at each time--attribute position:
\begin{equation}
\mathbf{z}_{b,t,k} = [\, x_{b,t,k}, \; m_{b,t,k} \,] \in \mathbb{R}^2.
\end{equation}
Each token is projected into a $D$-dimensional embedding space via a learnable linear map:
\begin{equation}
\mathbf{e}_{b,t,k} = \mathbf{W}_{\text{proj}} \mathbf{z}_{b,t,k} \in \mathbb{R}^D.
\end{equation}
The resulting tensor has shape $\mathbb{R}^{B \times L \times K \times D}$.

\paragraph{Query construction.}
To obtain a single representation per window, we reshape and average over all time--attribute embeddings to obtain a window-level query:
\begin{equation}
\mathbf{q}_b = \frac{1}{LK} \sum_{t=1}^{L} \sum_{k=1}^{K} \mathbf{e}_{b,t,k}
\;\;\in\; \mathbb{R}^{D}.
\end{equation}

This query captures the instance-specific characteristics of the partially observed window, including both observed values and missingness patterns. This is then used to retrieve relevant global patterns from the prototype bank, which encodes dataset-level temporal regimes.

\paragraph{Prototype bank and cross-attention.}
We maintain a learnable prototype bank $P=\{p_m\}_{m=1}^{M}$ with $p_m\in\mathbb{R}^{D}$, intended to capture characteristic temporal regimes (periodic behaviour, irregular fluctuations, or cross-channel correlations) that may not be fully observable within any single partially observed window. Multi-head cross-attention from the window query $q_b$ to the prototype bank yields a context vector $c_b\in\mathbb{R}^{D}$, which is mapped to the attribute domain by a learnable projection $W_r\in\mathbb{R}^{K\times D}$:
\begin{equation}
c_b=\mathrm{MHCrossAttn}(q_b;P),\qquad r_b=W_r c_b\in\mathbb{R}^{K}.
\end{equation}
The instance-dependent regime vector $r_b$ thus translates globally learned temporal regimes into window-level, attribute-specific signals used to condition the diffusion process. Full multi-head cross-attention details are provided in Appendix~\ref{appn:implementation}.

\subsection{Global-conditioned diffusion model}
We integrate the regime vector $r_b$ from Section~\ref{proto_module} into the diffusion-based imputation framework, refining the conditioning signal with instance-specific global information rather than relying on local observations or spectral features alone.

\subsubsection{Conditional diffusion formulation}
We follow the standard DDPM formulation~\cite{tashiro2021csdi}: the forward process gradually corrupts a clean input $x_0$ into $x_T$ via a Markov chain with a predefined noise schedule, admitting the closed-form $x_t=\sqrt{\bar\alpha_t}\,x_0+\sqrt{1-\bar\alpha_t}\,\epsilon$ with $\epsilon\sim\mathcal{N}(0,\mathbf{I})$. The reverse process $p_\theta(x_{t-1}\mid x_t)=\mathcal{N}(x_{t-1};\mu_\theta(x_t,t),\sigma_t^2\mathbf{I})$ is parameterised by a denoising network $\epsilon_\theta(x_t,t)$ trained to predict the injected noise $\epsilon$. For imputation, the reverse process is conditioned on a signal $h$ derived from the partially observed window: $p_\theta(x^{ta}_{0:T}\mid h)=p(x_T^{ta})\prod_t p_\theta(x^{ta}_{t-1}\mid x^{ta}_t,h)$, where $x_0^{ta}$ denotes the missing (target) entries. Existing approaches typically construct $h$ from local observations or window-level frequency representations; ProCTI refines this signal with prototype-derived global information.

\subsubsection{Theoretical perspective on local and global conditioning}\label{sec:theory}
Let $X\in\mathbb{R}^{L\times K}$ denote a multivariate time series window, with observed entries $O$ and missing entries $Y$; the imputation task is to model $p(Y\mid O)$. We formalise the distinction between local and global conditioning, then state four results characterising the value of hybrid conditioning during reverse diffusion.

\vspace{0.5em}
\noindent
\textbf{Definition 1 (Local conditioning).}
Local conditioning refers to information derived directly from the observed values in the input window. Formally, we define a local conditioning representation as
\[
h_{\text{loc}} = \phi_{\text{loc}}(O),
\]
where $\phi_{\text{loc}}$ is a function of observed values and missingness patterns within the current window.

\vspace{0.5em}
\noindent
\textbf{Definition 2 (Global conditioning).}
Global conditioning refers to information derived from dataset-level structure beyond the current input window. We define a latent regime variable $Z$ as an unobserved random variable representing global temporal patterns governing the data distribution, such as periodic structure, long-range trends, or cross-channel dependencies. In ProCTI, this is approximated by a learned prototype bank, yielding a representation
\[
h_{\text{glob}} = \phi_{\text{glob}}(O; P),
\]
where $P$ denotes the prototype bank learned over the training dataset.

\vspace{0.5em}
\noindent
\textbf{Definition 3 (Local-global conditioning).}
We define hybrid conditioning as the combination of local and global components:
\[
h = \psi(h_{\text{loc}}, h_{\text{glob}}).
\]

To implement hybrid conditioning, we build upon a frequency-aware diffusion framework~\cite{yang2024frequency} that decomposes window-level frequency content into a high-frequency component $h^{hf}\in\mathbb{R}^{B\times L\times K}$ and a dominant-frequency component $h^{dom}\in\mathbb{R}^{B\times L\times K}$. The dominant-frequency component captures coarse temporal structure such as periodicity and long-range trends — properties associated with dataset-level regimes — while the high-frequency component encodes fine-grained instance-specific variation. We therefore inject the regime vector $r_b\in\mathbb{R}^{K}$ from the prototype module into the dominant-frequency branch only:
\begin{equation}
h^{dom+r}_{b,t,k}=h^{dom}_{b,t,k}+\alpha r_{b,k},\qquad
h=\left[h^{hf}_{b,t};h^{dom+r}_{b,t}\right]\in\mathbb{R}^{2K},
\end{equation}
where $\alpha$ is a learnable scalar controlling the strength of global correction. The resulting prototype-refined conditioning $h$ is fed to the denoising network $\epsilon_\theta(x_t,t\mid h)$ at each diffusion step.

\paragraph{Theoretical results.}
We next state a sequence of results: standard uncertainty and score-based arguments, followed by a latent-regime characterisation of precisely when hybrid conditioning improves imputation. Lemma~1 records that conditioning on a global variable $Z$, when informative, reduces residual uncertainty about $Y$. Proposition~1 decomposes the conditional score that the denoising network approximates into a local term and a regime guidance term, providing a constructive interpretation of how global conditioning intervenes in reverse diffusion. Proposition~2 connects this decomposition to the denoising mean-squared error that the diffusion training objective directly minimises. Proposition~3 specialises to a latent-regime data model and gives equivalent conditions on $(Y,O,Z)$ under which the antecedents of the previous results hold, delineating both when ProCTI's prototype mechanism should be expected to help and when it should not.

\vspace{0.4em}
\noindent\textbf{Lemma 1 (Conditional uncertainty reduction).}
If $Z$ carries information about $Y$ beyond that already contained in $O$, namely $I(Y;Z\mid O)>0$, then $H(Y\mid O,Z)<H(Y\mid O).$

\noindent\textit{Proof.}
By the definition of conditional mutual information, $I(Y;Z\mid O)=H(Y\mid O)-H(Y\mid O,Z)$, so positivity of the left-hand side is equivalent to a strict decrease of conditional entropy. \hfill$\square$

The lemma states the information-theoretic premise that global information, when relevant, reduces residual uncertainty in $Y$, but does not by itself describe how this reduction is realised inside the reverse diffusion process. The next two propositions make that connection explicit, first at the level of the score function approximated by the denoising network, then at the level of the mean-squared error directly minimised by training.

\vspace{0.4em}
\noindent\textbf{Proposition 1 (Score decomposition under hybrid conditioning).}
Under hybrid conditioning $h=(O,Z)$, the conditional score of the noisy state $X_t$ admits the decomposition
\begin{equation}
\nabla_{x_t}\log p_t(x_t\mid O,Z)
=\underbrace{\nabla_{x_t}\log p_t(x_t\mid O)}_{\text{local score}}
+\underbrace{\nabla_{x_t}\log p(Z\mid x_t,O)}_{\text{regime guidance}}.
\label{eq:score-decomp}
\end{equation}

\noindent\textit{Proof.}
By Bayes' rule, $p_t(x_t\mid O,Z)=p(Z\mid x_t,O)\,p_t(x_t\mid O)/p(Z\mid O)$. Taking logarithms and differentiating in $x_t$, the term $\log p(Z\mid O)$ vanishes since it does not depend on $x_t$. \hfill$\square$

The decomposition has the same structure as classifier guidance for conditional diffusion~\cite{dhariwal2021diffusion,ho2022classifier} and assigns each conditioning source a distinct role: the local term is the score of the noised target given only $O$, while the regime guidance term pulls $x_t$ toward states compatible with $Z$. ProCTI does not have direct access to $Z$ at inference time; instead, its prototype-conditioning module learns an estimator $G=\phi_{\mathrm{glob}}(O;P)$ of regime-relevant information and feeds $G$ to the denoising network as a feasible surrogate for the regime guidance term in (\ref{eq:score-decomp}).

\vspace{0.4em}
\noindent\textbf{Proposition 2 (MMSE reduction under hybrid conditioning).}
Let $\mu^\star(x_t,h)=\mathbb{E}[X_0\mid X_t=x_t,h]$ denote the Bayes-optimal denoiser under conditioning $h$, and let $h_1=O$ and $h_2=(O,Z)$. Then
\begin{equation}
\mathbb{E}\!\left[\|X_0-\mu^\star(X_t,h_1)\|^2\right]-\mathbb{E}\!\left[\|X_0-\mu^\star(X_t,h_2)\|^2\right]
=\mathbb{E}\!\left[\bigl\|\mu^\star(X_t,h_2)-\mu^\star(X_t,h_1)\bigr\|^2\right]\ge 0,
\label{eq:mmse-reduction}
\end{equation}
with strict inequality whenever $Z$ carries information about $X_0$ beyond that already in $(X_t,O)$.

\noindent\textit{Proof.}
By the tower property, $\mu^\star(X_t,h_1)=\mathbb{E}[\mu^\star(X_t,h_2)\mid X_t,O]$. Decomposing $X_0-\mu^\star(X_t,h_1)=(X_0-\mu^\star(X_t,h_2))+(\mu^\star(X_t,h_2)-\mu^\star(X_t,h_1))$, the cross-term has zero expectation by the orthogonality of conditional means. Squaring and taking expectations yields the identity, with strict inequality precisely when the two optimal denoisers differ on a set of positive measure. \hfill$\square$

Proposition~2 is the operational counterpart of Lemma~1: it records the corresponding reduction in mean-squared error of the optimal denoiser, which is the quantity directly minimised by the diffusion training objective. Together, Propositions~1 and~2 give a unified picture: global conditioning enters the score additively as a regime guidance term, and this addition can only decrease the optimal denoising error, strictly so whenever the guidance term is informative.

\vspace{0.4em}
\noindent\textbf{Latent-regime data model.}
The previous three results all depend on the antecedent that $Z$ carries information about $Y$ (or $X_0$) beyond $O$. To pin down when this antecedent holds, we adopt a latent-regime model in which each window is associated with $Z\in\{1,\ldots,M^{\star}\}$ with $(O,Y)\mid Z\sim p(\cdot,\cdot\mid Z)$. The model is non-restrictive; any joint distribution over $X$ admits such a representation and serves only to make the conditions for prototype-based improvement explicit.

\vspace{0.4em}
\noindent\textbf{Proposition 3 (Conditions for prototype informativeness).}
Under the latent-regime data model, the following are equivalent:
\begin{enumerate}[label=(\roman*),noitemsep,topsep=2pt]
\item $I(Y;Z\mid O)>0$;
\item $Y\not\!\perp\!\!\!\perp Z\mid O$;
\item $\inf_{f}\mathbb{E}[\|Y-f(O)\|^2]>\inf_{g}\mathbb{E}[\|Y-g(O,Z)\|^2]$.
\end{enumerate}
When any of these conditions hold, Lemma~1 yields a strict reduction in conditional entropy, and Proposition~2 yields a strict reduction in optimal denoising mean-squared error from hybrid over local conditioning. When they fail, that is when $Y\!\perp\!\!\!\perp Z\mid O$, no function of $Z$ adds information about $Y$ beyond $O$, and prototype-derived global conditioning cannot improve the optimal denoiser.

\noindent\textit{Proof.}
$(i)\Leftrightarrow(ii)$ is the standard equivalence between vanishing conditional mutual information and conditional independence. For $(ii)\Leftrightarrow(iii)$, the optimal predictor of $Y$ given $(O,Z)$ in mean-squared error is $\mathbb{E}[Y\mid O,Z]$, which coincides almost surely with $\mathbb{E}[Y\mid O]$ if and only if $Y\!\perp\!\!\!\perp Z\mid O$. \hfill$\square$

Proposition~3 delineates the regime in which ProCTI's prototype-conditioning mechanism is expected to provide gains. When $Y$ is well-approximated by a deterministic function of $O$ — as when an entire missing channel can be recovered by linear regression on the remaining channels — $Y\!\perp\!\!\!\perp Z\mid O$ holds approximately, and no global conditioning signal can improve on a sufficiently expressive local denoiser. We return to this prediction in Section~\ref{sec:experiments}, where it explains the dataset-dependence of ProCTI's gains observed under attribute-wise missingness. In the complementary regime, where $Z$ carries residual information about $Y$ given $O$, the prototype-conditioning module aims to recover this information from observations alone: under the latent-regime model with bank size $M\ge M^{\star}$ and a sufficiently expressive attention map ($\pi_m(O)\approx p(Z=m\mid O)$), the retrieved regime vector $r_b$ functions as a soft sufficient statistic for $Z$ given $O$, and Propositions~1 and~2 quantify the corresponding score and MMSE improvements.

%% file: 4_experiments_v3.tex
\section{Experiments}\label{sec:experiments}

This section describes the experiments used to evaluate ProCTI against state-of-the-art baseline methods in terms of imputation effectiveness and robustness under attribute-wise missing scenarios. All experiments were conducted on a machine with an Intel Core i7-13700K CPU, two NVIDIA GeForce RTX 4090 GPUs (24GB each), and 128GB RAM.

\subsection{Experimental Setup}
\textbf{Datasets} We used five real-world time series datasets for our experiments, comparable to other similar work~\cite{yuan2024diffusion, zhou2024mtsci, yang2024frequency, du2024tsi}. \textit{PhysioNet} \cite{reyna2020early} comprises multivariate clinical time-series data collected from ICU patients, with each record comprising measurements of approximately 40 physiological variables. The \textit{Beijing} \cite{beijing_multi-site_air_quality_501} dataset includes air-pollutant data collected from 12 air-quality monitoring sites in Beijing. The \textit{Weather} \cite{jena_climate} dataset is from the weather station of the Max Planck Institute for Biogeochemistry, comprising 14 meteorological attributes recorded between January 1st 2009 and December 31st 2016. The \textit{Stock} \cite{google_stock_yahoo} dataset reflects Google stock price data from 2004 to 2019, with each record representing a new day. Finally, the \textit{Gait} \cite{ngo2014largest} dataset comprises gait data collected using IMU sensors from 744 users during walks on level ground. We provide further details about dataset preparation for our experiments in Appendix~\ref{appn:datasetprep}.

\textbf{Baselines} We evaluate ProCTI against a diverse set of baselines. We consider imputation methods including BRITS~\cite{cao2018brits}, CSDI~\cite{tashiro2021csdi}, MTSCI~\cite{zhou2024mtsci}, FGTI~\cite{yang2024frequency} and TIDER~\cite{liu2023multivariate}. We include two forecasting models, SCINet~\cite{liu2022scinet} and iTransformer~\cite{liu2023itransformer}, that can be adapted to imputation via masked-value reconstruction~\cite{du2024tsi}. We also evaluate two diffusion-based generation models, Diffusion-TS~\cite{yuan2024diffusion} and PaD-TS~\cite{li2025population}, that can likewise be adapted for conditional imputation. 

\textbf{Evaluation metrics} For all of our experiments, we report the MAE and RMSE values to reflect imputation quality in terms of closeness to the ground truth. Therefore, the lower the value, the better the imputation. While MAE reflects the average error, RMSE reflects extreme deviations from the ground truth. 
\begin{table*}[t]
\centering
\scriptsize
\caption{Comparison of imputation effectiveness across baseline methods, aggregated over missingness ratios $\{0.1,0.3,0.5,0.7\}$. Each cell reports the mean with the standard deviation as a subscript. The best mean MAE and RMSE are shown in bold, and the second-best results are underlined. The full table of average results for each missingness ratio is in Appendix~\ref{appn:results}.}
\begin{adjustbox}{width=\textwidth}
\begin{tabular}{ll|cccccccccc}
\toprule
Dataset & Metric & BRITS & CSDI & SCINet & TIDER & MTSCI & DiffusionTS & iTransformer & FGTI & PaD-TS & ProCTI \\
\midrule

\multirow{2}{*}{Gait}
& MAE  & $0.352_{\pm.025}$ & $0.449_{\pm.017}$ & $0.475_{\pm.040}$ & $0.746_{\pm.002}$ & $0.501_{\pm.019}$ & $0.367_{\pm.009}$ & $0.537_{\pm.035}$ & $\underline{0.332}_{\pm.026}$ & $0.942_{\pm.032}$ & $\textbf{0.313}_{\pm.026}$ \\
& RMSE & $\underline{0.533}_{\pm.037}$ & $0.726_{\pm.014}$ & $0.671_{\pm.054}$ & $0.985_{\pm.003}$ & $0.750_{\pm.019}$ & $0.585_{\pm.014}$ & $0.765_{\pm.039}$ & $0.563_{\pm.035}$ & $1.197_{\pm.032}$ & $\textbf{0.518}_{\pm.041}$ \\
\midrule

\multirow{2}{*}{PhysioNet}
& MAE  & $0.505_{\pm.032}$ & $0.456_{\pm.034}$ & $0.401_{\pm.091}$ & $0.751_{\pm.030}$ & $0.526_{\pm.075}$ & $0.544_{\pm.026}$ & $0.413_{\pm.098}$ & $\underline{0.362}_{\pm.082}$ & $0.865_{\pm.062}$ & $\textbf{0.259}_{\pm.058}$ \\
& RMSE & $3.401_{\pm.158}$ & $4.919_{\pm.203}$ & $\underline{1.557}_{\pm.443}$ & $3.465_{\pm.558}$ & $1.609_{\pm.210}$ & $3.054_{\pm.698}$ & $1.925_{\pm.509}$ & $2.970_{\pm.431}$ & $3.153_{\pm.551}$ & $\textbf{1.383}_{\pm.299}$ \\
\midrule

\multirow{2}{*}{Weather}
& MAE  & $0.148_{\pm.046}$ & $0.125_{\pm.015}$ & $0.276_{\pm.085}$ & $0.649_{\pm.006}$ & $0.278_{\pm.115}$ & $0.246_{\pm.037}$ & $0.283_{\pm.116}$ & $\underline{0.113}_{\pm.049}$ & $0.176_{\pm.013}$ & $\textbf{0.095}_{\pm.037}$ \\
& RMSE & $0.355_{\pm.040}$ & $0.334_{\pm.020}$ & $0.434_{\pm.094}$ & $0.833_{\pm.008}$ & $0.472_{\pm.119}$ & $0.493_{\pm.056}$ & $0.442_{\pm.111}$ & $\underline{0.298}_{\pm.034}$ & $0.362_{\pm.018}$ & $\textbf{0.274}_{\pm.040}$ \\
\midrule

\multirow{2}{*}{Beijing}
& MAE  & $0.217_{\pm.073}$ & $0.186_{\pm.091}$ & $0.341_{\pm.070}$ & $0.722_{\pm.001}$ & $0.282_{\pm.126}$ & $0.256_{\pm.028}$ & $0.405_{\pm.123}$ & $\underline{0.159}_{\pm.053}$ & $0.699_{\pm.073}$ & $\textbf{0.153}_{\pm.048}$ \\
& RMSE & $0.482_{\pm.076}$ & $0.464_{\pm.119}$ & $0.559_{\pm.085}$ & $1.003_{\pm.003}$ & $0.577_{\pm.137}$ & $0.582_{\pm.022}$ & $0.660_{\pm.138}$ & $\underline{0.437}_{\pm.070}$ & $0.988_{\pm.079}$ & $\textbf{0.414}_{\pm.070}$ \\
\midrule

\multirow{2}{*}{Stock}
& MAE  & $3.880_{\pm.245}$ & $2.842_{\pm.926}$ & $2.671_{\pm.812}$ & $5.278_{\pm.049}$ & $3.762_{\pm.494}$ & $4.302_{\pm.363}$ & $\underline{2.622}_{\pm.971}$ & $3.186_{\pm.422}$ & $4.987_{\pm.150}$ & $\textbf{2.223}_{\pm.660}$ \\
& RMSE & $4.299_{\pm.248}$ & $3.223_{\pm.976}$ & $3.482_{\pm.919}$ & $5.712_{\pm.039}$ & $4.120_{\pm.488}$ & $4.801_{\pm.240}$ & $\underline{3.226}_{\pm.963}$ & $4.046_{\pm.230}$ & $5.531_{\pm.089}$ & $\textbf{2.808}_{\pm.666}$ \\
\bottomrule
\end{tabular}
\end{adjustbox}
\label{tab:markovmask_results_aggregated}
\end{table*}

\subsection{Imputation effectiveness}
We compare imputation effectiveness under stochastic segment masking, artificially masking observed values to simulate missing data. We refer to the resulting proportion of unavailable values as the missing ratio \(r\). Specifically, we implement the masking strategy of Zerveas et al.~\cite{zerveas2021transformer} as a two-state Markov process that generates alternating masked and unmasked segments whose lengths follow geometric distributions. The process is parameterised by a target missing ratio \(r\) and an average masked segment length \(l_m\), with the corresponding average unmasked segment length given by \(l_u=\frac{1-r}{r}l_m\). During training, we set \(r=0.15\), while evaluation is conducted at missing ratios of 0.10, 0.30, 0.50, and 0.70, where larger ratios correspond to increasingly challenging imputation settings. For both training and evaluation, we set \(l_m=6\), as moderately longer contiguous gaps provide a more challenging missingness pattern than short isolated gaps, while preserving sufficient observed context for reconstruction. Table~\ref{tab:markovmask_results_aggregated} reports the mean and standard deviation of MAE and RMSE values over five random runs for each missingness ratio. We provide the full table of average results for each missingness ratio in Table~\ref{tab:markovmask_results} in Appendix~\ref{appn:results}. 

Overall, ProCTI outperforms all baselines in terms of average performance. This demonstrates the benefit of combining global prototype priors with local conditioning. Among the baselines, FGTI attains the second-best performance, yielding the second-lowest mean MAE for all scenarios and the second-lowest mean RMSE in 6 out of 10 scenarios. In comparison to FGTI, ProCTI reduces the average MAE by approximately $3$--$30\%$. The largest gains are observed on PhysioNet and Stock, followed by Weather, Beijing and Gait.

\subsection{Robustness under attribute-wise missingness}\label{sec:attribute-wisemissing}
\begin{table*}[t]
\centering
\scriptsize
\caption{Structured missingness evaluation results. Best results are in bold and second-best are underlined.}
\begin{adjustbox}{width=\textwidth}
\begin{tabular}{lll|cccccccccc}
\toprule
Dataset & Metric & Drop & BRITS & CSDI & SCINet & TIDER & MTSCI & DiffusionTS & iTransformer & FGTI & PaD-TS & ProCTI \\
\midrule

\multirow{4}{*}{Gait}
& \multirow{2}{*}{MAE}
& 1 & 0.735 & 0.801 & 0.745 & 0.745 & 0.741 & 1.053 & 0.745 & \textbf{0.616} & 1.037 & \underline{0.619} \\
& & 2 & 0.718 & 0.808 & 0.748 & 0.748 & 0.745 & 1.059 & 0.748 & \underline{0.643} & 1.044 & \textbf{0.640} \\

& \multirow{2}{*}{RMSE}
& 1 & 1.017 & 1.039 & 0.984 & 0.985 & 0.961 & 1.311 & 0.984 & \underline{0.839} & 1.293 & \textbf{0.821} \\
& & 2 & 0.971 & 1.047 & 0.988 & 0.988 & 0.967 & 1.316 & 0.988 & \underline{0.877} & 1.300 & \textbf{0.851} \\

\midrule
\multirow{4}{*}{PhysioNet}
& \multirow{2}{*}{MAE}
& 1 & 0.931 & 0.841 & 0.836 & \underline{0.772} & 0.946 & 0.780 & 0.805 & 0.856 & 0.980 & \textbf{0.747} \\
& & 2 & 0.962 & 0.694 & 0.708 & 0.768 & 0.740 & 0.693 & 0.692 & \underline{0.645} & 0.944 & \textbf{0.619} \\

& \multirow{2}{*}{RMSE}
& 1 & 1.246 & 1.010 & 1.031 & \underline{0.938} & 1.147 & 1.002 & 0.998 & 1.042 & 1.215 & \textbf{0.903} \\
& & 2 & 1.281 & 0.890 & 0.885 & 0.969 & 0.922 & 0.929 & 0.874 & \underline{0.831} & 1.183 & \textbf{0.781} \\

\midrule

\multirow{4}{*}{Weather}
& \multirow{2}{*}{MAE}
& 1 & \textbf{0.205} & 0.687 & 0.669 & 0.664 & 0.667 & 0.671 & 0.664 & 0.310 & 0.724 & \underline{0.269} \\
& & 2 & \textbf{0.212} & 0.693 & 0.679 & 0.674 & 0.671 & 0.696 & 0.673 & 0.318 & 0.755 & \underline{0.288} \\

& \multirow{2}{*}{RMSE}
& 1 & \textbf{0.438} & 0.859 & 0.840 & 0.838 & 0.830 & 0.874 & 0.838 & 0.593 & 0.932 & \underline{0.479} \\
& & 2 & \textbf{0.480} & 0.872 & 0.855 & 0.852 & 0.848 & 0.902 & 0.852 & 0.561 & 0.969 & \underline{0.494} \\

\midrule

\multirow{4}{*}{Beijing}
& \multirow{2}{*}{MAE}
& 1 & 1.107 & 0.708 & 0.737 & 0.745 & 0.647 & 0.875 & 0.736 & \underline{0.429} & 0.863 & \textbf{0.419} \\
& & 2 & 0.927 & 0.700 & 0.736 & 0.734 & 0.653 & 0.881 & 0.735 & \underline{0.426} & 0.877 & \textbf{0.436} \\

& \multirow{2}{*}{RMSE}
& 1 & 1.484 & 0.981 & 1.053 & 1.030 & 0.915 & 1.180 & 1.052 & \underline{0.769} & 1.173 & \textbf{0.716} \\
& & 2 & 1.309 & 0.990 & 1.053 & 1.094 & 0.937 & 1.191 & 1.053 & \underline{0.776} & 1.189 & \textbf{0.746} \\

\midrule

\multirow{4}{*}{Stock}
& \multirow{2}{*}{MAE}
& 1 & \textbf{2.177} & 4.975 & 5.350 & 5.073 & 4.882 & 4.655 & 5.092 & 5.422 & 4.776 & \underline{4.580} \\
& & 2 & \textbf{2.355} & 5.079 & 5.367 & 5.146 & 5.374 & 4.827 & 5.166 & 5.391 & 4.800 & \underline{4.714} \\

& \multirow{2}{*}{RMSE}
& 1 & \textbf{2.528} & 5.522 & 5.797 & 5.516 & 5.347 & 5.152 & 5.533 & 5.784 & 5.243 & \underline{4.995} \\
& & 2 & \textbf{2.689} & 5.581 & 5.778 & 5.550 & 5.793 & 5.282 & 5.566 & 5.787 & 5.308 & \underline{5.082} \\

\bottomrule
\end{tabular}
\end{adjustbox}
\label{tab:dropexperiments}
\end{table*}
We implemented attribute-wise missingness scenarios to evaluate robustness under structured missing patterns. This setting is motivated by real-world situations in which one or more attributes are missing for a period of time, for example, due to sensor malfunctions or system outages. Using the same training configuration as in the previous experiments, we conducted two attribute-wise missing protocols: Drop 1 and Drop 2, where one attribute and two attributes, respectively, were randomly selected and fully masked within the test windows. During evaluation, the reverse diffusion process reconstructed the missing attribute(s) solely from the remaining observed variables. For all datasets, we considered all available attributes except PhysioNet, which exhibited high sparsity. For this dataset, only attributes with more than 80\% observed values were included in the experiment. Table~\ref{tab:dropexperiments} reports the MAE and RMSE values on the masked attributes for the different methods.

\begin{figure*}[t]
\centering

\begin{subfigure}[t]{0.32\textwidth}
    \centering
    \includegraphics[width=\textwidth]{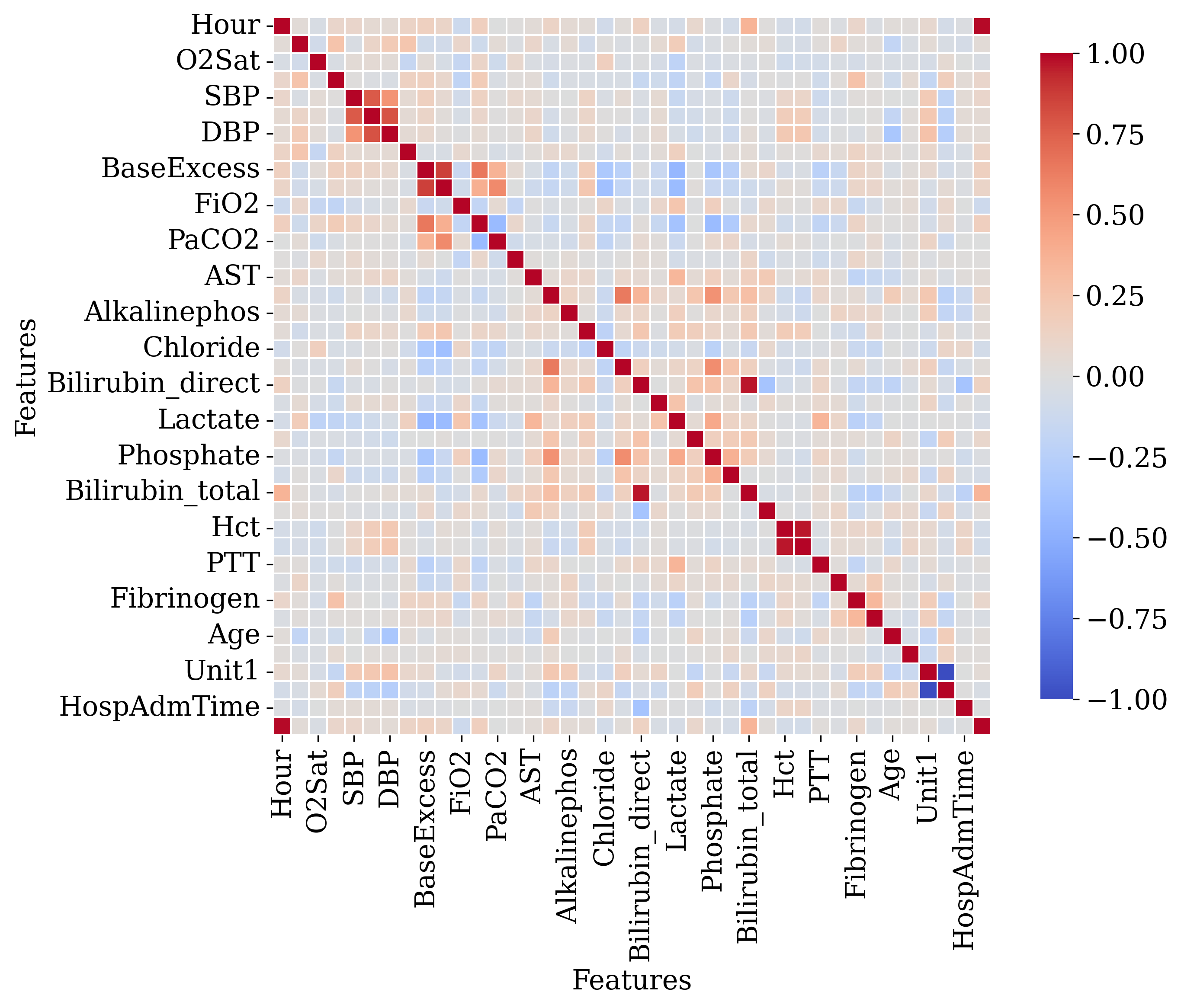}
    \caption{PhysioNet}
    \label{fig:corr_physionet}
\end{subfigure}
\hfill
\begin{subfigure}[t]{0.32\textwidth}
    \centering
    \includegraphics[width=\textwidth]{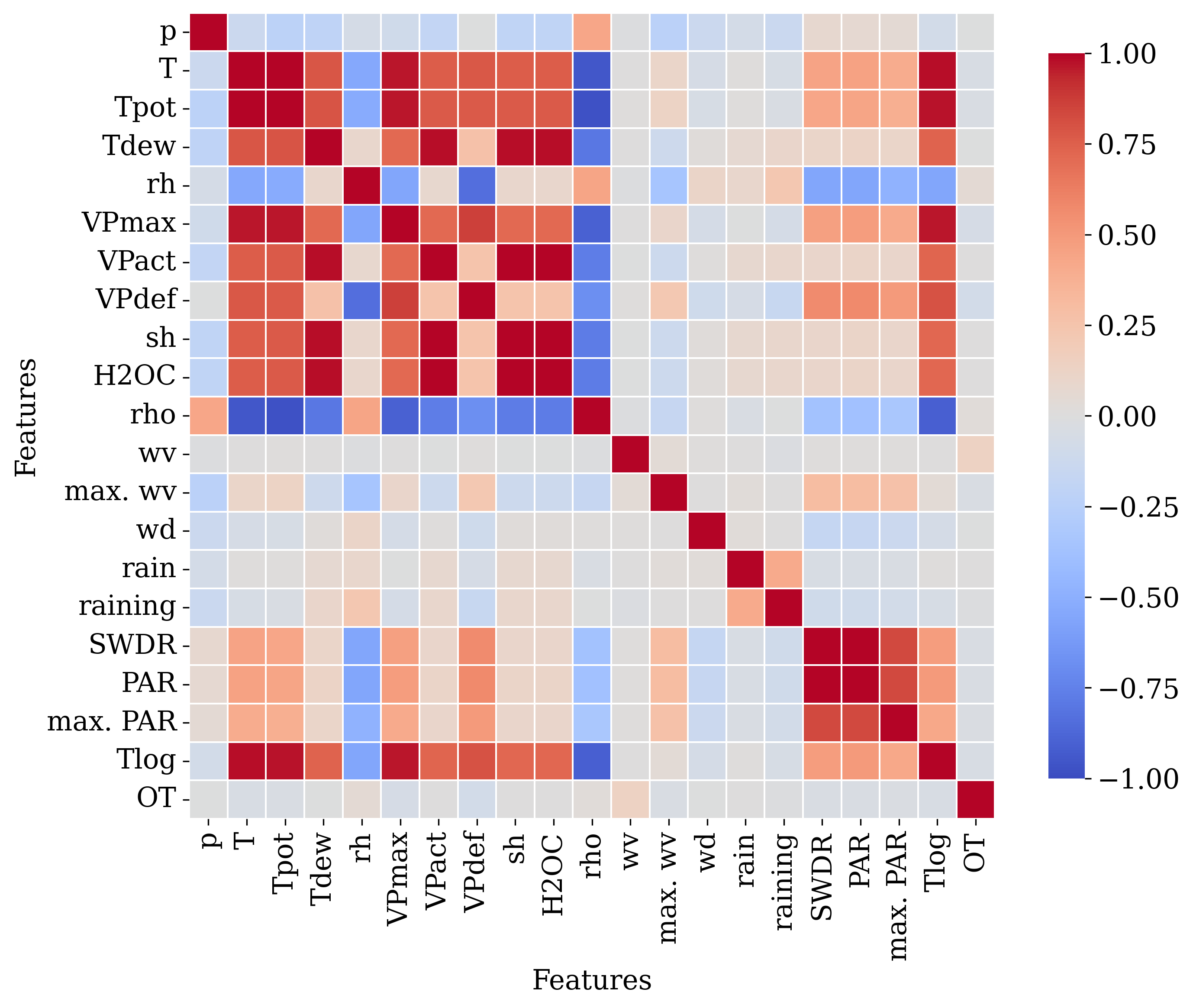}
    \caption{Weather}
    \label{fig:corr_weather}
\end{subfigure}
\hfill
\begin{subfigure}[t]{0.32\textwidth}
    \centering
    \includegraphics[width=\textwidth]{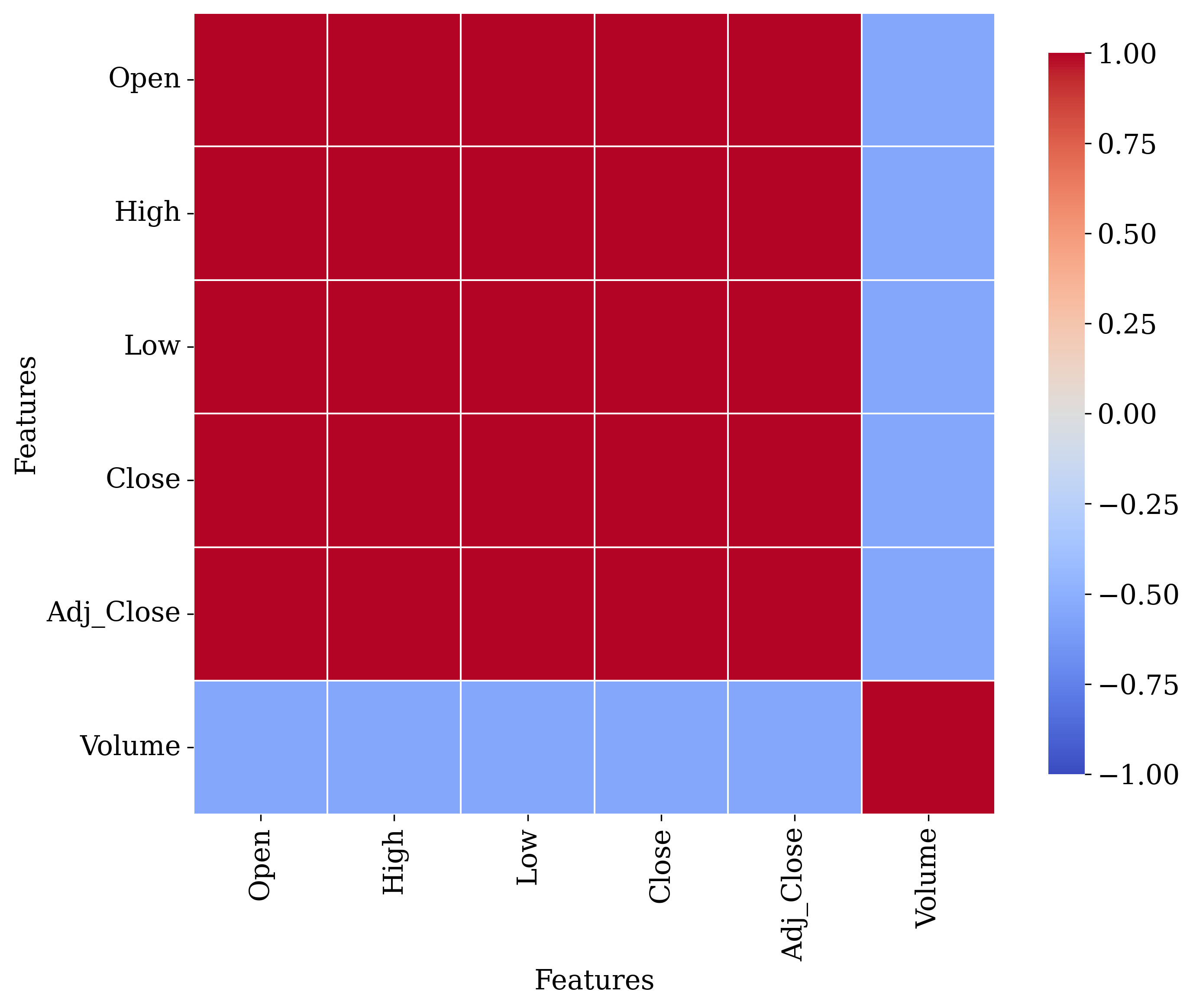}
    \caption{Stock}
    \label{fig:corr_stock}
\end{subfigure}

\vspace{0.8em}

\vspace{0.8em}

\begin{center}
\begin{subfigure}[t]{0.31\textwidth}
    \centering
    \includegraphics[width=\textwidth]{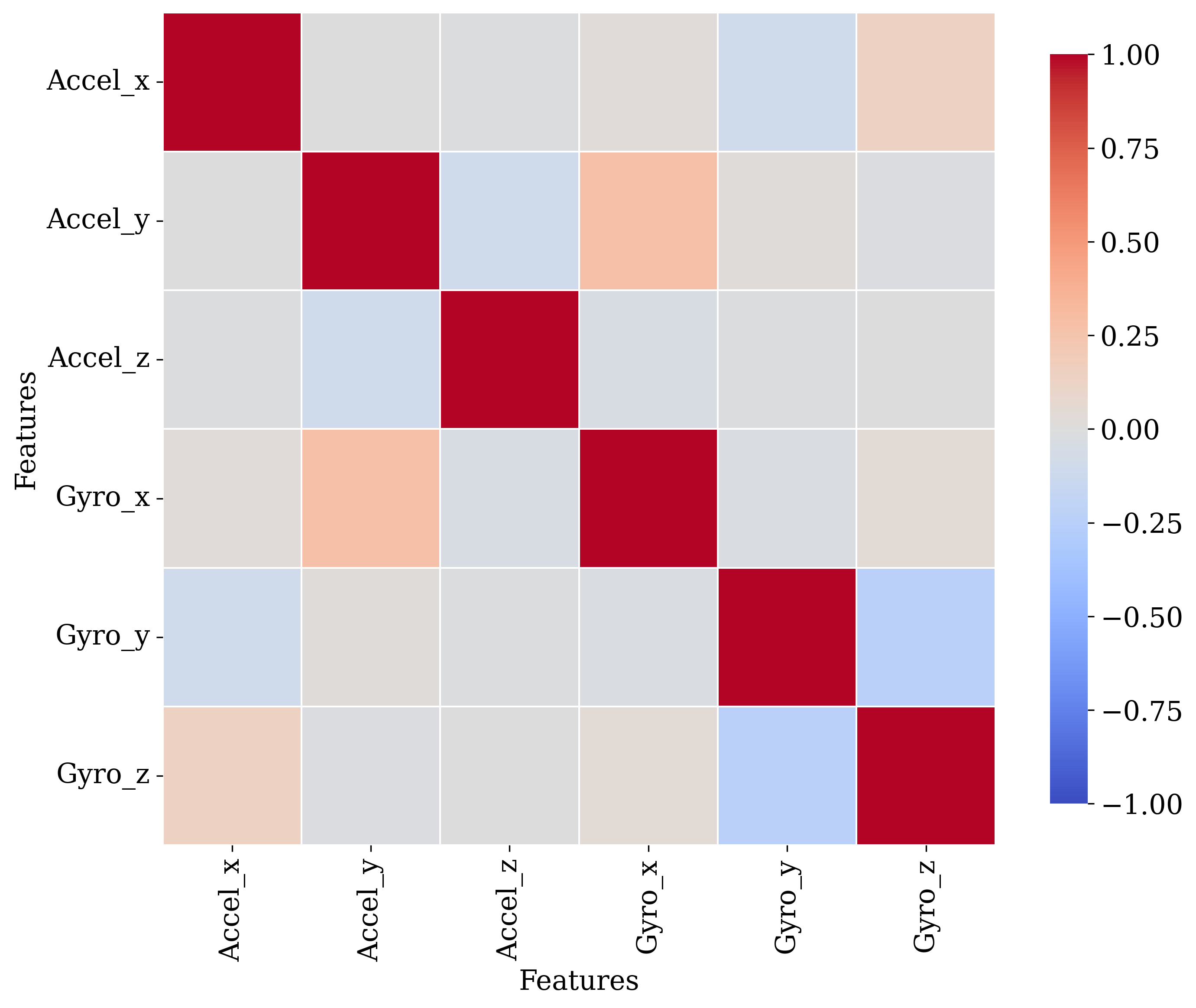}
    \caption{Gait}
    \label{fig:corr_gait}
\end{subfigure}
\hspace{0.04\textwidth}
\begin{subfigure}[t]{0.31\textwidth}
    \centering
    \includegraphics[width=\textwidth]{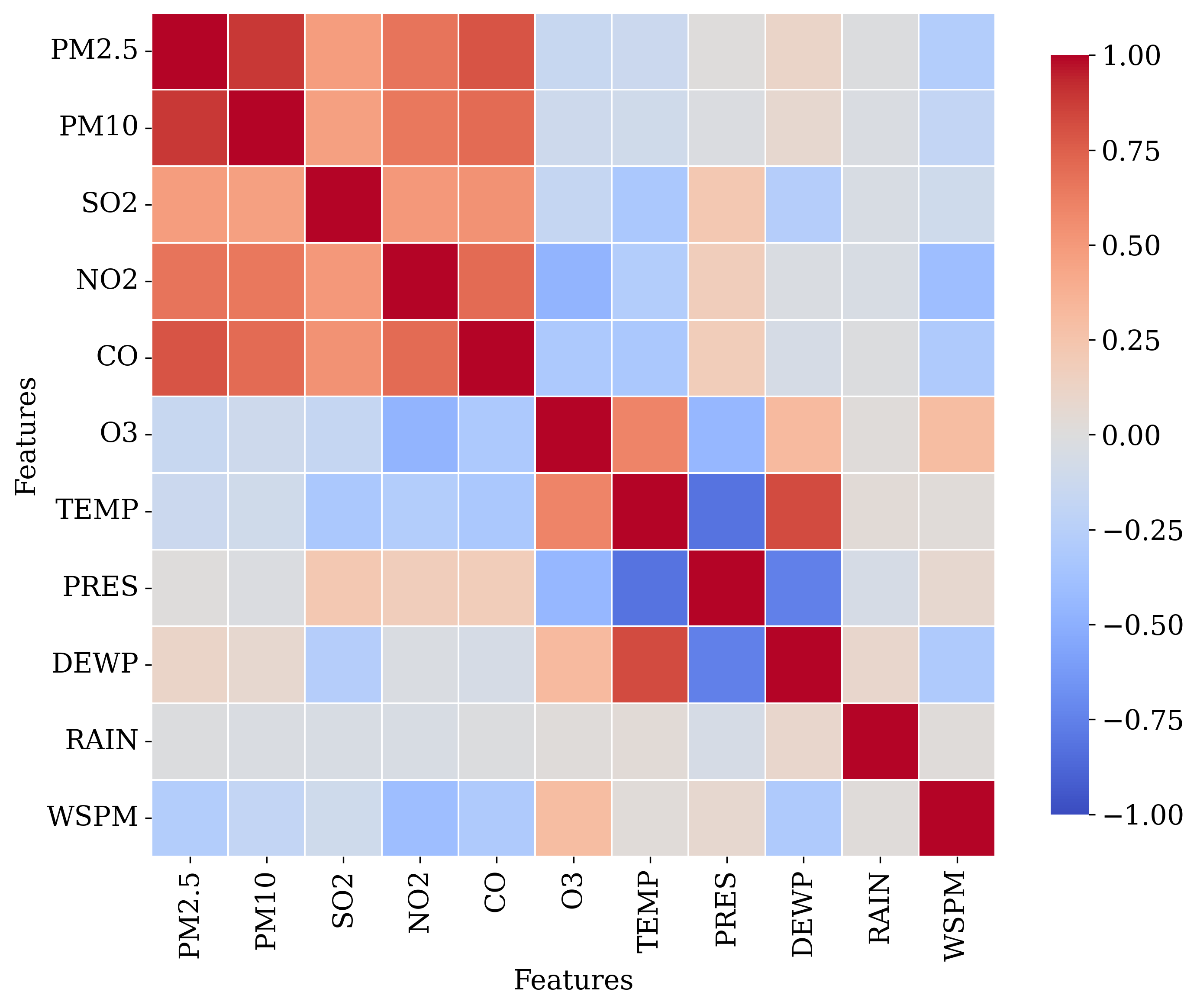}
    \caption{Beijing}
    \label{fig:corr_beijing}
\end{subfigure}
\end{center}

\caption{Feature correlation heatmaps for the five datasets: (a) PhysioNet, (b) Weather, (c) Stock, (d) Gait, and (e) Beijing.}
\label{fig:correlation_heatmaps}
\end{figure*}

For the Gait, PhysioNet, and Beijing datasets, ProCTI outperforms all competing methods in terms of overall performance, achieving the lowest MAE and RMSE values in the majority of attribute-drop scenarios. FGTI consistently emerges as the second-best performing method. For the Weather and Stock datasets, however, BRITS displays the lowest MAE and RMSE values. Inspecting the feature-wise correlations of the five datasets, we find that BRITS’s superior performance in these two cases can be attributed to its explicit feature-regression mechanism combined with bidirectional temporal recurrence. These are particularly effective when there is strong cross-feature redundancy. In the Stock dataset, for example, price variables such as Open, High, Low, Close, and Adj\_Close are almost perfectly correlated, while many variables in the Weather dataset also exhibit linear dependencies and tightly coupled feature groups. Under attribute-drop evaluation, where an entire variable is missing, these relationships allow BRITS to reconstruct the missing channel directly from the remaining observed variables, consequently reducing the relative advantage of more complex diffusion-based or forecasting-adapted models. This pattern matches Proposition~3 in Section~\ref{sec:theory}: the strong cross-feature redundancy in Stock and Weather approximately satisfies the conditional independence $Y\!\perp\!\!\!\perp Z\mid O$, the regime in which prototype-derived global conditioning provably cannot improve upon a sufficiently expressive local predictor. We provide heatmap visualisations of the datasets’ feature-wise correlations in Figure~\ref{fig:correlation_heatmaps}. 

In contrast, the Gait and PhysioNet datasets display substantially weaker feature correlations, indicating that missing-attribute recovery cannot rely primarily on simple cross-variable redundancy. For the Beijing dataset, although some strong pairwise correlations are present, its dependency structure is considerably more heterogeneous, containing mixtures of strongly positive, strongly negative, and weak relationships across pollutant, meteorological, and dispersion-related variables. Furthermore, as illustrated in Figure~\ref{fig:intro_regimes}, Beijing windows form multiple recurring operating regimes with distinct feature profiles, suggesting that accurate missing-channel reconstruction depends on broader contextual interactions rather than uniform redundancy alone. Under such conditions, the antecedent of Proposition~3 holds: regimes carry residual information about $Y$ beyond $O$, and ProCTI's prototype-derived global conditioning enables it to outperform BRITS and the remaining baselines, consistent with the entropy and mean-squared-error reductions established by Lemma~1 and Proposition~2.

\subsection{Ablation analysis}

We study the influence of the main components introduced in ProCTI, namely the prototype bank, the cross-attention interaction between input windows and prototypes, and the scaling parameter $\alpha$ used to inject regime information. We also study the effect of injecting regime information into both dominant and high-frequency branches together, as well as separately. To isolate the contribution of each component, we consider seven variants: (A) the original FGTI baseline without prototype conditioning, (B) adding only a learnable prototype bank, (C) enabling prototype cross-attention, (D) using a frozen prototype bank with attention, and (E - G) the full ProCTI model with a learnable $\alpha$ injected into the dominant frequency branch (inj-dom), high frequency branch (inj-hf), and both branches (inj-both) respectively. Full numerical results and discussion are reported in Table~\ref{tab:ablation_appn}.

Across all five datasets, introducing a learnable prototype bank (Variant B) consistently improves upon the FGTI baseline (Variant A), highlighting the benefit of incorporating global prototype priors in addition to local frequency-based conditioning. Enabling prototype cross-attention (Variant C) yields further gains on most datasets, suggesting that dynamically matching each input window to relevant prototypes is more effective than relying on static prototype representations alone. In contrast, Variant D shows degraded performance relative to the preceding variants, indicating that a learnable prototype bank is more effective than a frozen one. Comparing the results across Variants E, F and G reflects that injecting a learnable $\alpha$ into the dominant frequency branch only yields lower error (with the exception of PhysioNet where Variant F returns a lower MAE and RMSE). Overall, Variant E yields the best results in the majority of cases, demonstrating the value of learning $\alpha$ (allowing the model to regulate how strongly regime information influences the local conditioning signal depending on the dataset) and injecting the regime information into the dominant frequency branch only. 

\begin{table*}[t]
\centering
\scriptsize
\caption{Ablation study of ProCTI across all datasets (mean values only). Variant A denotes the base FGTI model without prototype conditioning. Variants B--D use fixed $\alpha=0.1$, while Variant E uses a learnable $\alpha$.}
\setlength{\tabcolsep}{4pt}
\begin{tabular}{c|ccc|cc|cc|cc|cc|cc}
\toprule
\multirow{2}{*}{\textbf{Variant}} &
\multirow{2}{*}{\makecell{\textbf{Proto}\\\textbf{Bank}}} &
\multirow{2}{*}{\makecell{\textbf{Proto}\\\textbf{Attn}}} &
\multirow{2}{*}{\makecell{\textbf{Learnable}\\$\boldsymbol{\alpha}$}} &
\multicolumn{2}{c}{\textbf{Gait}} &
\multicolumn{2}{c}{\textbf{PhysioNet}} &
\multicolumn{2}{c}{\textbf{Weather}} &
\multicolumn{2}{c}{\textbf{Beijing}} &
\multicolumn{2}{c}{\textbf{Stock}} \\
\cmidrule(lr){5-6}
\cmidrule(lr){7-8}
\cmidrule(lr){9-10}
\cmidrule(lr){11-12}
\cmidrule(lr){13-14}
 & & & & MAE & RMSE & MAE & RMSE & MAE & RMSE & MAE & RMSE & MAE & RMSE \\
\midrule
A & -- & -- & --
& 0.322 & 0.531
& 0.228 & 0.460
& 0.149 & 0.336
& 0.1566 & 0.4338
& 3.2606 & 4.0410 \\
B & \checkmark & -- & --
& 0.318 & 0.521
& 0.219 & 0.424
& 0.113 & 0.296
& 0.1518 & 0.4062
& 2.6317 & 3.2982 \\
C & \checkmark & \checkmark & --
& 0.317 & 0.520
& 0.217 & 0.417
& 0.119 & 0.304
& 0.1519 & 0.4044
& 2.4705 & 3.0971 \\
D & frozen & \checkmark & --
& 0.316 & 0.521
& 0.221 & 0.434
& 0.139 & 0.329
& 0.1591 & 0.4362
& 2.9831 & 3.7186 \\
E (inj-dom) & \checkmark & \checkmark & \checkmark
& \textbf{0.316} & \textbf{0.518}
& 0.219 & 0.414
& \textbf{0.096} & \textbf{0.280}
& \textbf{0.1515} & \textbf{0.4043}
& \textbf{2.2686} & \textbf{2.8323} \\
F (inj-hf) & \checkmark & \checkmark & \checkmark
& 0.320 & 0.520
& \textbf{0.214} & \textbf{0.412}
& 0.102 & 0.288
& 0.1910 & 0.4746
& 3.1198 & 3.7978 \\
G (inj-both) & \checkmark & \checkmark & \checkmark
& 0.316 & 0.526
& 0.223 & 0.422
& 0.104 & 0.289
& 0.1938 & 0.4752
& 2.8556 & 3.4772 \\
\bottomrule
\end{tabular}
\label{tab:ablation_appn}
\end{table*}

%% file: 5_conclusion_v3.tex
\section{Conclusion}
In this paper, we introduced ProCTI, a diffusion-based imputation framework that combines local observations with reusable global priors through prototype retrieval. The proposed framework incorporates a learnable prototype-conditioning module that provides global context at the window-level, augmenting the conditioning signals of the reverse diffusion process. We further present a theoretical framework that distinguishes local and global conditioning in diffusion-based imputation models and demonstrate how ProCTI combines both, thereby improving imputation quality. Extensive experiments across diverse real-world datasets demonstrate that ProCTI consistently outperforms strong baselines under both random missingness and attribute-wise missingness settings. The gains of prototype-derived conditioning are conditional on regime structure being informative beyond observed entries (Proposition~3): when $Y\!\perp\!\!\!\perp Z\mid O$ holds approximately, as under strong cross-feature redundancy, simpler regression baselines remain competitive. Future work includes prototype-bank interpretability and designs for non-stationary regimes.

%% file: appendix_v3.tex
\appendix
\section{Appendix}
\subsection{Related work}\label{appn:relatedwork}
\textbf{Prototype-based representations in time series modelling} Prototype-based representations have appeared across the broader time-series literature. In anomaly detection, ProtoAD~\cite{li2023prototypes} uses prototypes to characterise regular latent patterns and improve interpretability through example-based explanations. In healthcare imputation, PRIME~\cite{yu2025imputation} leverages prototype memory to capture inter-series information from similar patients within irregular clinical records. In generation, recent works such as TimeDP~\cite{huang2025timedp}, OATS~\cite{deng2026oats}, and K-ProtoDiff~\cite{duan2026k} employ prototypes as domain prompts, guidance signals, or key-pattern constraints for synthesising realistic sequences. In contrast, ProCTI introduces prototypes for \emph{diffusion-based conditional imputation} of general multivariate time series, where prototypes are learned as reusable dataset-level regimes and explicitly combined with local observations during reverse diffusion. This differs from prior prototype methods that focus on retrieval-style inter-series memory, interpretability, or unconditional generation rather than principled global conditioning for imputation.

\subsection{ProCTI implementation details}\label{appn:implementation}

\subsubsection{Multi-head cross-attention details}
We provide here the full multi-head cross-attention computation summarised in Section~\ref{proto_module}. Let the number of heads be $H$, with per-head dimension $d_h=D/H$. For each head $h\in\{1,\dots,H\}$ we have learnable projections $W_Q^{(h)},W_K^{(h)},W_V^{(h)}\in\mathbb{R}^{d_h\times D}$, and the projected query, key, and value vectors are
\[
q_b^{(h)}=W_Q^{(h)}q_b,\quad
k_m^{(h)}=W_K^{(h)}p_m,\quad
v_m^{(h)}=W_V^{(h)}p_m,
\]
with $q_b^{(h)},k_m^{(h)},v_m^{(h)}\in\mathbb{R}^{d_h}$. The attention weights are
\[
\pi_{b,m}^{(h)}=\frac{\exp\!\left((q_b^{(h)})^\top k_m^{(h)}/\sqrt{d_h}\right)}{\sum_{j=1}^{M}\exp\!\left((q_b^{(h)})^\top k_j^{(h)}/\sqrt{d_h}\right)},
\]
yielding the per-head context vector $c_b^{(h)}=\sum_{m=1}^{M}\pi_{b,m}^{(h)}v_m^{(h)}\in\mathbb{R}^{d_h}$. The outputs of all heads are concatenated and projected to obtain $c_b=W_O\,\mathrm{Concat}\!\left(c_b^{(1)},\dots,c_b^{(H)}\right)\in\mathbb{R}^{D}$ with $W_O\in\mathbb{R}^{D\times D}$. The regime vector is then $r_b=W_r c_b\in\mathbb{R}^{K}$ as in Section~\ref{proto_module}.

\subsubsection{Training and imputation algorithms}
We provide the training and imputation algorithms for our proposed model ProCTI, in Algorithm~\ref{alg:procti_training} and Algorithm~\ref{alg:procti_imputation} respectively.

\begin{algorithm}[h]
\caption{Training process of ProCTI}
\label{alg:procti_training}
\begin{algorithmic}[1]

\Input Time series $X$, keep-mask $M$, prototype bank $P$, diffusion steps $T$
\Output Trained $\epsilon_\theta(\cdot)$ and prototype module

\Repeat
    \State $\hat{X}_0 \leftarrow X \odot (1-M)$, \quad $X^C \leftarrow X \odot M$
    \State $h^{hf}, h^{dom} \leftarrow \mathrm{FreqFilter}(X^C)$
    \State $q_b \leftarrow \mathrm{Query}(X^C,M)$
    \State $c_b \leftarrow \mathrm{CrossAttn}(q_b,P)$, \quad $r_b \leftarrow W_r c_b$
    \State $h^{dom+r}_{b,t,k} \leftarrow h^{dom}_{b,t,k}+\alpha r_{b,k}$
    \State $t \sim \mathrm{Uniform}(\{1,\dots,T\})$, \quad $\epsilon \sim \mathcal{N}(0,I)$
    \State $\hat{X}_t \leftarrow \sqrt{\bar{\alpha}_t}\hat{X}_0+\sqrt{1-\bar{\alpha}_t}\epsilon$
    \State $\hat{\epsilon} \leftarrow \epsilon_\theta(\hat{X}_t,t \mid X^C,h^{hf},h^{dom+r})$
    \State Update parameters using $\mathcal{L}_{\mathrm{diff}}=\|\epsilon-\hat{\epsilon}\|_2^2$
\Until{converged}

\end{algorithmic}
\end{algorithm}

\begin{algorithm}[h]
\caption{Imputation process of ProCTI}
\label{alg:procti_imputation}
\begin{algorithmic}[1]

\Input Incomplete time series $X$, keep-mask $M$, trained prototype bank $P$, trained denoising network $\epsilon_\theta(\cdot)$, diffusion steps $T$
\Output Imputed time series $\tilde{X}$

\State $X^C \leftarrow X \odot M$
\State $h^{hf}, h^{dom} \leftarrow \mathrm{FreqFilter}(X^C)$
\State $h^{dom+r} \leftarrow \mathrm{ProtoCond}(X^C,M,P,h^{dom})$
\State $\hat{X}_T \sim \mathcal{N}(0,I)$

\For{$t=T,T-1,\dots,1$}
    \State $\hat{\epsilon} \leftarrow \epsilon_\theta(\hat{X}_t,t \mid X^C,h^{hf},h^{dom+r})$
    \State Obtain $\hat{X}_{t-1}$ from $p_\theta(\hat{X}_{t-1}\mid \hat{X}_t,X^C,h^{hf},h^{dom+r})$
\EndFor

\State $\tilde{X} \leftarrow X \odot M + \hat{X}_0 \odot (1-M)$
\State \Return $\tilde{X}$

\end{algorithmic}
\end{algorithm}

\subsection{Dataset preparation}\label{appn:datasetprep}
In this section, we provide a detailed description of the dataset preprocessing steps to create training, validation and test splits for our evaluation experiments. Since the Gait, PhysioNet, and Beijing datasets contain identifiable entities that can be used for partitioning (user IDs, patient IDs, and monitoring stations, respectively), we construct training, validation, and test splits such that each entity appears in only one split, preventing information leakage. This strategy differs from conventional approaches that randomly partition data at the window level, and is therefore designed to evaluate the ability of ProCTI and baseline methods to generalise across non-overlapping entities by capturing global patterns. 

Specifically, for the Gait and PhysioNet datasets, we adopt user-disjoint and patient-wise splits, respectively, ensuring that all sequences associated with a given subject or patient are confined to a single partition. The training, validation, and test splits follow a 70:15:15 ratio at the entity level. For the Beijing dataset, we use a station-disjoint split, partitioning the 12 monitoring stations into training, validation, and test groups in an 8:2:2 ratio, and constructing windows independently within each station. For the Weather and Stock datasets, we adopt a chronological split along the temporal dimension, using a 70:15:15 ratio for training, validation, and testing. In all cases, non-overlapping fixed-length windowing is applied after splitting to generate sequences for model training and evaluation. 

\subsection{Supplementary experiments}
\subsubsection{Probabilistic evaluation}
\begin{table*}[h]
\centering
\scriptsize
\setlength{\tabcolsep}{4pt}
\caption{Probabilistic evaluation on the test split across missingness ratios on PhysioNet data.}
\label{tab:crps_results_physionet}
\begin{tabular}{cc|cccccc}
\toprule
\textbf{Metric} & \textbf{Missing} &
\textbf{CSDI} &
\textbf{Diffusion-TS} &
\textbf{FGTI} &
\textbf{PaD-TS} &
\textbf{MTSCI} &
\textbf{ProCTI} \\
\midrule

\multirow{4}{*}{CRPS$(\downarrow)$}
& 0.10 & 0.1798 & 0.6579 & 0.1187 & 0.8727 & 0.4076 & 0.1286 \\
& 0.30 & 0.1867 & 0.8361 & 0.1362 & 0.8955 & 0.5980 & 0.1390 \\
& 0.50 & 0.2223 & 0.8485 & 0.1647 & 1.0370 & 0.6081 & 0.1593 \\
& 0.70 & 0.2284 & 0.7599 & 0.2266 & 0.9718 & 0.7307 & 0.2040 \\
\midrule

\multirow{4}{*}{PI90 Coverage$(~90\%)$}
& 0.10 & 0.8447 & 0.0368 & 0.8479 & 0.1383 & 0.8053 & 0.8467 \\
& 0.30 & 0.8460 & 0.0514 & 0.9097 & 0.2291 & 0.7806 & 0.8876 \\
& 0.50 & 0.8566 & 0.0734 & 0.9307 & 0.3375 & 0.7612 & 0.9017 \\
& 0.70 & 0.8666 & 0.1082 & 0.9265 & 0.4910 & 0.7368 & 0.8872 \\
\midrule

\multirow{4}{*}{PI90 Width$(\downarrow)$}
& 0.10 & 0.5861 & 0.0647 & 0.5884 & 0.2509 & 1.2602 & 0.5704 \\
& 0.30 & 0.6368 & 0.0806 & 0.7693 & 0.4349 & 1.4165 & 0.7012 \\
& 0.50 & 0.7264 & 1.1117 & 1.0723 & 0.6777 & 1.5938 & 0.9137 \\
& 0.70 & 0.8134 & 0.1625 & 1.5036 & 1.0513 & 1.7813 & 1.1878 \\
\bottomrule
\end{tabular}
\end{table*}

\begin{table*}[t]
\centering
\scriptsize
\setlength{\tabcolsep}{4pt}
\caption{Probabilistic evaluation on the test split across missingness ratios on Beijing data.}
\label{tab:crps_results_beijing}
\begin{tabular}{cc|cccccc}
\toprule
\textbf{Metric} & \textbf{Missing} &
\textbf{CSDI} &
\textbf{Diffusion-TS} &
\textbf{FGTI} &
\textbf{MTSCI} &
\textbf{PaD-TS} &
\textbf{ProCTI} \\
\midrule

\multirow{4}{*}{CRPS$(\downarrow)$}
& 0.10 & 0.1646 & 0.2368 & 0.1204 & 0.2286 & 0.6141 & 0.1185 \\
& 0.30 & 0.1700 & 0.2326 & 0.1282 & 0.2374 & 0.5920 & 0.1254 \\
& 0.50 & 0.1894 & 0.2393 & 0.1472 & 0.2639 & 0.5784 & 0.1424 \\
& 0.70 & 0.2119 & 0.2400 & 0.1679 & 0.2861 & 0.5657 & 0.1617 \\
\midrule

\multirow{4}{*}{PI90 Coverage$(~90\%)$}
& 0.10 & 0.8439 & 0.3016 & 0.8152 & 0.7506 & 0.1611 & 0.8536 \\
& 0.30 & 0.8479 & 0.3478 & 0.8279 & 0.7657 & 0.2946 & 0.8748 \\
& 0.50 & 0.8536 & 0.3950 & 0.8353 & 0.7825 & 0.4410 & 0.8925 \\
& 0.70 & 0.8679 & 0.4417 & 0.8301 & 0.8177 & 0.5870 & 0.9057 \\
\midrule

\multirow{4}{*}{PI90 Width$(\downarrow)$}
& 0.10 & 0.8081 & 0.1771 & 0.5441 & 0.8911 & 0.2970 & 0.6532 \\
& 0.30 & 0.8539 & 0.2057 & 0.5857 & 1.0115 & 0.5531 & 0.7235 \\
& 0.50 & 0.9440 & 0.2409 & 0.6544 & 1.1976 & 0.8731 & 0.8427 \\
& 0.70 & 1.1174 & 0.2789 & 0.7461 & 1.4426 & 1.2630 & 1.0122 \\
\bottomrule
\end{tabular}
\end{table*}

To complement the imputation evaluation reported in the main experiments (Section~\ref{sec:experiments}), we further assess the probabilistic quality of the predictive distributions produced by diffusion-based models. Using the PhysioNet and Beijing datasets under the Markov masking protocol, we adopt the Continuous Ranked Probability Score (CRPS)~\cite{tashiro2021csdi,yang2024frequency}, and 90\% prediction interval coverage (PI90 coverage) and width (PI90 width)~\cite{stankeviciute2021conformal} across different missingness ratios. CRPS evaluates the overall quality of the predictive distribution in comparison to the ground truth, where lower values are preferred. PI90 coverage measures how often the ground-truth values fall within the predicted 90\% interval. Values closer to 0.90 are preferred, meaning that well-calibrated predictive intervals are neither too narrow nor excessively wide. Finally, PI90 width reflects interval sharpness, where narrower intervals are preferred, provided calibration remains close to the nominal 90\% level. Table~\ref{tab:crps_results_physionet} and Table~\ref{tab:crps_results_beijing} show the results for PhysioNet and Beijing, respectively.

Overall ProCTI consistently achieves strong probabilistic performance across all missingness ratios, obtaining the best or second-best CRPS values while maintaining PI90 Coverage close to 90\%. In contrast, some baselines such as Diffusion-TS and PaD-TS, produce narrow intervals but reveal severe under-coverage, indicating overconfident uncertainty estimates. Although FGTI achieves the lowest CRPS in some settings under PhysioNet, it generally requires wider prediction intervals to do so. Overall, ProCTI provides the most balanced trade-off between predictive accuracy, calibration, and sharpness, demonstrating that incorporating global prototype priors improves imputation performance as well as probabilistic reliability.

\subsubsection{Resource consumption}\label{appn:resource}

\begin{figure*}[h]
    \centering
    \includegraphics[width=0.86\textwidth]{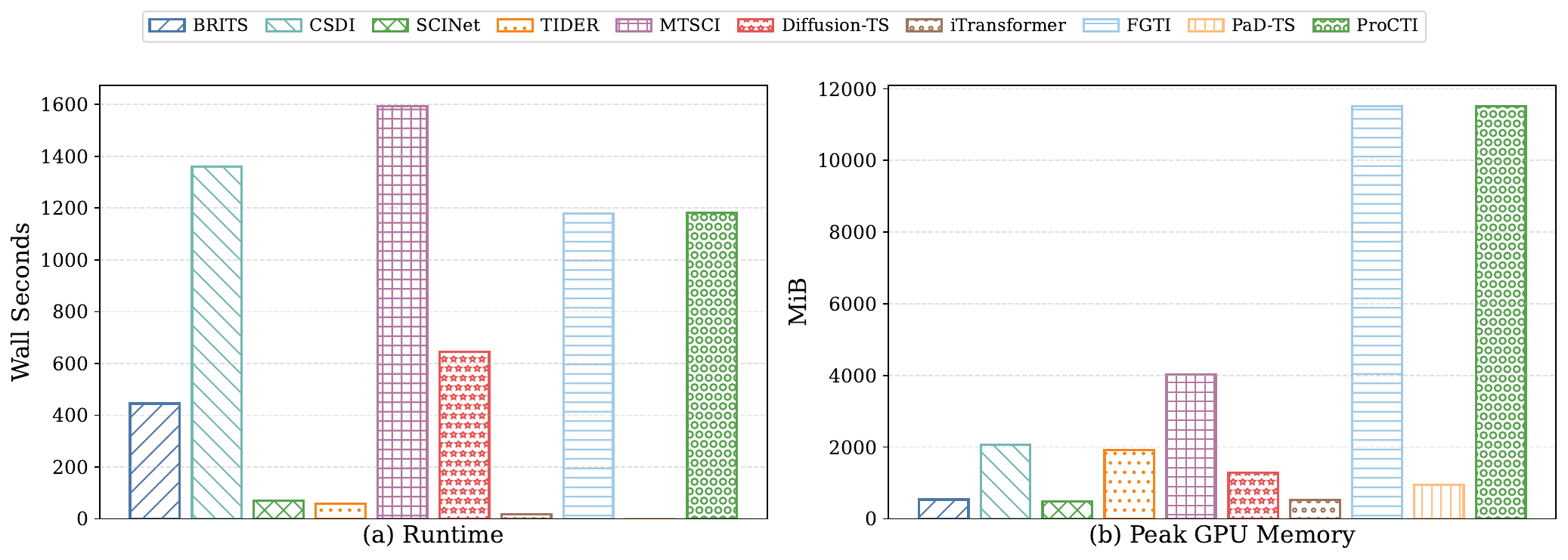}
    \caption{Runtime and peak GPU memory usage for different models on the Beijing dataset under the 30\% missingness test setting.}
    \label{fig:beijing_resource_usage}
\end{figure*}

We evaluate the resource consumption of ProCTI against all baselines in Figure~\ref{fig:beijing_resource_usage}. ProCTI achieves runtime comparable to FGTI and lower than other diffusion-based methods such as CSDI and MTSCI. Although its GPU memory usage is higher than most baselines, it remains similar to FGTI despite the added model complexity due to the prototype conditioning module. Given its overall superior imputation performance, we argue that this additional resource cost is justified.

\subsubsection{Hyperparameter evaluation}\label{appn:hyperparameter}
\begin{figure}[h]
    \centering

    \includegraphics[width=0.7\linewidth]{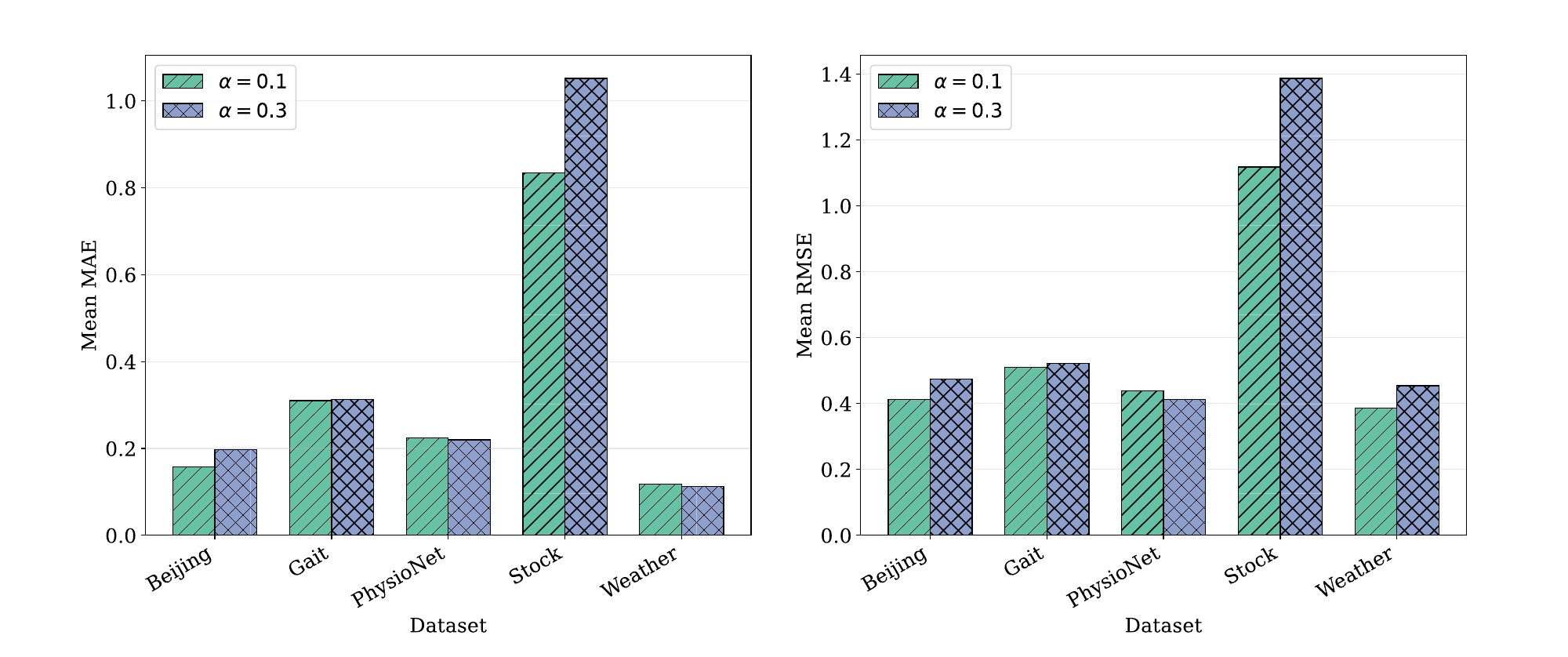}

    \caption{Effect of scaling parameter $\alpha \in \{0.1,0.3\}$ on imputation performance. Lower values indicate better imputation performance.}
    \label{fig:alpha_hyperparam}
\end{figure}

In this section, we present a hyperparameter sensitivity analysis investigating the effect of prototype bank size and the regime-vector scaling parameter $\alpha$. Table~\ref{tab:proto_bank_size} and Figure~\ref{fig:alpha_hyperparam} report ProCTI's imputation performance on held-out validation data, averaged across missingness ratios (10\%, 30\%, 50\%, and 70\%), for different values of these hyperparameters. Table~\ref{tab:proto_bank_size} shows that a prototype bank size of 32 yields lower error in the majority of cases across MAE and RMSE (6/10 cells). Figure~\ref{fig:alpha_hyperparam} indicates that $\alpha=0.1$ wins on Beijing, Gait, and Stock, while $\alpha=0.3$ gives lower MAE on PhysioNet and Weather. We therefore use $\alpha=0.1$ as it gives the best overall performance, and a prototype bank size of 32 in ProCTI.

\begin{table}[h]
\centering
\scriptsize
\setlength{\tabcolsep}{5pt}
\caption{Effect of prototype bank size (number of prototypes) on imputation performance. Best results for each dataset and metric are highlighted in bold.}
\label{tab:proto_bank_size}
\begin{tabular}{cc|ccccc}
\toprule
\textbf{Metric} & \textbf{Prototype bank size} & \textbf{Beijing} & \textbf{Gait} & \textbf{PhysioNet} & \textbf{Stock} & \textbf{Weather} \\
\midrule
\multirow{3}{*}{MAE}
& 16 & 0.200 & 0.310 & 0.222 & 0.894 & 0.102 \\
& 32 & \textbf{0.158} & 0.310 & 0.224 & \textbf{0.835} & 0.118 \\
& 64 & 0.202 & \textbf{0.306} & \textbf{0.216} & 1.030 & \textbf{0.095} \\
\midrule
\multirow{3}{*}{RMSE}
& 16 & 0.481 & 0.519 & 0.416 & 1.187 & 0.450 \\
& 32 & \textbf{0.412} & \textbf{0.510} & 0.438 & \textbf{1.118} & \textbf{0.387} \\
& 64 & 0.482 & 0.512 & \textbf{0.413} & 1.340 & 0.434 \\
\bottomrule
\end{tabular}
\end{table}


\subsection{Supplementary results} \label{appn:results}

\subsubsection{Imputation effectiveness}
In Table~\ref{tab:markovmask_results}, we report the detailed imputation performance of all methods under the Markov masking protocol, using MAE and RMSE at each missingness ratio, averaged over five repeated runs. Overall, ProCTI achieves the lowest errors in the majority of evaluated scenarios, while FGTI consistently delivers the second-best performance in most cases.
\begin{table*}[h]
\centering
\scriptsize
\caption{Comparison of imputation effectiveness across baseline methods. Best results are shown in bold and second-best results are underlined.}
\begin{adjustbox}{width=\textwidth}
\begin{tabular}{lll|cccccccccc}
\toprule
Dataset & Metric & Missing & BRITS & CSDI & SCINet & TIDER & MTSCI & DiffusionTS & iTransformer & FGTI & PaD-TS & ProCTI \\
\midrule

\multirow{8}{*}{Gait}
& \multirow{4}{*}{MAE}
& 0.1 & 0.324 & 0.436 & 0.436 & 0.749 & 0.484 & 0.358 & 0.501 & \underline{0.307} & 0.906 & \textbf{0.288} \\
& & 0.3 & 0.342 & 0.438 & 0.454 & 0.744 & 0.488 & 0.361 & 0.521 & \underline{0.317} & 0.928 & \textbf{0.299} \\
& & 0.5 & 0.360 & 0.449 & 0.482 & 0.745 & 0.505 & 0.369 & 0.544 & \underline{0.336} & 0.954 & \textbf{0.317} \\
& & 0.7 & 0.383 & 0.473 & 0.528 & 0.745 & 0.527 & 0.380 & 0.583 & \underline{0.366} & 0.981 & \textbf{0.347} \\

& \multirow{4}{*}{RMSE}
& 0.1 & \underline{0.491} & 0.717 & 0.615 & 0.989 & 0.734 & 0.573 & 0.722 & 0.530 & 1.162 & \textbf{0.477} \\
& & 0.3 & \underline{0.518} & 0.715 & 0.645 & 0.982 & 0.737 & 0.575 & 0.749 & 0.543 & 1.184 & \textbf{0.497} \\
& & 0.5 & \underline{0.545} & 0.726 & 0.682 & 0.984 & 0.755 & 0.588 & 0.774 & 0.570 & 1.208 & \textbf{0.528} \\
& & 0.7 & \underline{0.577} & 0.746 & 0.741 & 0.984 & 0.775 & 0.604 & 0.814 & 0.610 & 1.235 & \textbf{0.572} \\

\midrule

\multirow{8}{*}{PhysioNet}
& \multirow{4}{*}{MAE}
& 0.1 & 0.486 & 0.426 & 0.327 & 0.740 & 0.456 & 0.514 & 0.324 & \underline{0.273} & 0.794 & \textbf{0.203} \\
& & 0.3 & 0.481 & 0.427 & 0.339 & 0.796 & 0.473 & 0.574 & 0.353 & \underline{0.322} & 0.837 & \textbf{0.223} \\
& & 0.5 & 0.504 & 0.480 & 0.416 & 0.730 & 0.556 & 0.555 & 0.432 & \underline{0.397} & 0.897 & \textbf{0.278} \\
& & 0.7 & 0.551 & 0.489 & 0.524 & 0.737 & 0.617 & 0.534 & 0.543 & \underline{0.459} & 0.934 & \textbf{0.332} \\

& \multirow{4}{*}{RMSE}
& 0.1 & 3.245 & 4.758 & \underline{1.355} & 2.996 & 1.868 & 2.023 & 1.484 & 2.446 & 2.404 & \textbf{1.182} \\
& & 0.3 & 3.372 & 4.789 & \underline{1.083} & 4.274 & 1.355 & 3.538 & 1.540 & 2.841 & 3.076 & \textbf{1.074} \\
& & 0.5 & 3.366 & 5.203 & 1.679 & 3.264 & \textbf{1.623} & 3.413 & 2.120 & 3.459 & 3.609 & \underline{1.657} \\
& & 0.7 & 3.622 & 4.928 & 2.111 & 3.324 & \textbf{1.590} & 3.242 & 2.554 & 3.135 & 3.524 & \underline{1.621} \\

\midrule

\multirow{8}{*}{Weather}
& \multirow{4}{*}{MAE}
& 0.1 & 0.105 & 0.109 & 0.199 & 0.641 & 0.160 & 0.297 & 0.173 & \underline{0.073} & 0.163 & \textbf{0.062} \\
& & 0.3 & 0.124 & 0.120 & 0.226 & 0.651 & 0.221 & 0.249 & 0.215 & \underline{0.085} & 0.169 & \textbf{0.076} \\
& & 0.5 & 0.155 & 0.125 & 0.289 & 0.653 & 0.307 & 0.227 & 0.309 & \underline{0.111} & 0.177 & \textbf{0.097} \\
& & 0.7 & 0.209 & \textbf{0.144} & 0.389 & 0.651 & 0.425 & 0.213 & 0.435 & 0.182 & 0.193 & \underline{0.146} \\

& \multirow{4}{*}{RMSE}
& 0.1 & 0.315 & 0.309 & 0.343 & 0.821 & 0.338 & 0.570 & 0.346 & \underline{0.295} & 0.338 & \textbf{0.230} \\
& & 0.3 & 0.337 & 0.333 & 0.382 & 0.838 & 0.424 & 0.498 & 0.369 & \underline{0.268} & 0.362 & \textbf{0.266} \\
& & 0.5 & 0.357 & 0.335 & 0.452 & 0.837 & 0.508 & 0.465 & 0.462 & \underline{0.281} & 0.368 & \textbf{0.276} \\
& & 0.7 & 0.410 & 0.359 & 0.558 & 0.836 & 0.616 & 0.441 & 0.591 & \underline{0.346} & 0.379 & \textbf{0.326} \\

\midrule

\multirow{8}{*}{Beijing}
& \multirow{4}{*}{MAE}
& 0.1 & 0.141 & \textbf{0.096} & 0.332 & 0.721 & 0.152 & 0.225 & 0.285 & 0.105 & 0.619 & \underline{0.103} \\
& & 0.3 & 0.177 & \underline{0.130} & 0.267 & 0.722 & 0.212 & 0.243 & 0.334 & \underline{0.130} & 0.667 & \textbf{0.128} \\
& & 0.5 & 0.243 & 0.217 & 0.331 & 0.722 & 0.331 & 0.268 & 0.439 & \underline{0.176} & 0.720 & \textbf{0.169} \\
& & 0.7 & 0.306 & 0.299 & 0.435 & 0.722 & 0.435 & 0.288 & 0.562 & \underline{0.225} & 0.790 & \textbf{0.214} \\

& \multirow{4}{*}{RMSE}
& 0.1 & 0.397 & \underline{0.340} & 0.512 & 0.999 & 0.419 & 0.554 & 0.515 & 0.363 & 0.899 & \textbf{0.337} \\
& & 0.3 & 0.449 & \underline{0.400} & 0.485 & 1.007 & 0.519 & 0.578 & 0.590 & 0.406 & 0.959 & \textbf{0.381} \\
& & 0.5 & 0.508 & 0.508 & 0.562 & 1.004 & 0.636 & 0.589 & 0.701 & \underline{0.452} & 1.011 & \textbf{0.438} \\
& & 0.7 & 0.574 & 0.609 & 0.677 & 1.003 & 0.733 & 0.606 & 0.832 & \underline{0.526} & 1.084 & \textbf{0.499} \\

\midrule

\multirow{8}{*}{Stock}
& \multirow{4}{*}{MAE}
& 0.1 & 3.582 & 1.875 & 1.729 & 5.278 & 3.144 & 3.810 & \underline{1.520} & 2.678 & 4.773 & \textbf{1.348} \\
& & 0.3 & 3.801 & 2.355 & 2.382 & 5.212 & 3.626 & 4.684 & \underline{2.228} & 3.039 & 4.992 & \textbf{2.121} \\
& & 0.5 & 3.983 & 3.151 & \underline{2.931} & 5.293 & 3.988 & 4.340 & 2.960 & 3.370 & 5.073 & \textbf{2.546} \\
& & 0.7 & 4.153 & 3.986 & \underline{3.640} & 5.328 & 4.292 & 4.373 & 3.781 & 3.658 & 5.108 & \textbf{2.876} \\

& \multirow{4}{*}{RMSE}
& 0.1 & 4.010 & 2.198 & 2.387 & 5.712 & 3.532 & 4.492 & \underline{2.099} & 3.736 & 5.413 & \textbf{1.913} \\
& & 0.3 & 4.203 & \textbf{2.715} & 3.174 & 5.660 & 3.946 & 5.076 & 2.861 & 4.018 & 5.517 & \underline{2.720} \\
& & 0.5 & 4.399 & \underline{3.557} & 3.826 & 5.728 & 4.344 & 4.795 & 3.607 & 4.171 & 5.572 & \textbf{3.162} \\
& & 0.7 & 4.584 & 4.423 & 4.540 & 5.750 & 4.658 & 4.843 & \underline{4.338} & 4.260 & 5.620 & \textbf{3.438} \\

\bottomrule
\end{tabular}
\end{adjustbox}
\label{tab:markovmask_results}
\end{table*}

\subsubsection{Statistical significance of imputation effectiveness}

As the imputation performance of ProCTI and FGTI were very similar on some datasets such as Gait and Beijing, we statistically analysed the significance of the gains yielded by ProCTI. We conducted paired t-tests and Wilcoxon signed-rank tests across the 5 seeds for each missing ratio, with the one-sided hypothesis that ProCTI's per-seed error is lower than FGTI's (i.e., FGTI - ProCTI > 0). The paired t-test assumes the per-seed differences are approximately normally distributed, while the Wilcoxon signed-rank test is a non-parametric alternative that does not require this assumption and is more robust with small sample sizes. Results are shown in Table~\ref{tab:significance} (significance at the 0.05 shown in bold).

On Beijing, ProCTI significantly outperforms FGTI (p < 0.05) in 4/8 cells (t-test) and 1/8 (Wilcoxon). On Gait, ProCTI is significantly better in 5/8 cells under both tests, concentrated at $r \in {0.1, 0.3, 0.5, 0.7}$; significance is lost at r = 0.7, consistent with the smaller gains in Table 1. These tests are based on 5 seeds which is a small sample for reliably estimating normality or the sign/rank distribution. This likely explains why some cells miss significance despite ProCTI's lower mean error in Table 1. We expect significance to hold more consistently with additional seeds. 

\begin{table}[h!]
\centering
\scriptsize
\caption{Paired one-sided significance tests for ProCTI vs.\ FGTI across 5 seeds ($H_1$: ProCTI error $<$ FGTI error). $p$-values below 0.05 are in bold. With $n=5$, the minimum attainable Wilcoxon $p$-value is $1/32 \approx 0.031$.}
\begin{tabular}{lll|cccc}
\toprule
Dataset & $r$ & Metric & $t$-stat & $t$-test $p$ & Wilcoxon stat & Wilcoxon $p$ \\
\midrule
\multirow{8}{*}{Beijing}
& \multirow{2}{*}{0.1}
  & MAE  & -0.306 & 0.388 & 5 & 0.313 \\
& & RMSE & -2.590 & \textbf{0.030} & 1 & 0.063 \\
\cmidrule(lr){2-7}
& \multirow{2}{*}{0.3}
  & MAE  & -3.111 & \textbf{0.018} & 0 & \textbf{0.031} \\
& & RMSE & -2.156 & \textbf{0.049} & 2 & 0.094 \\
\cmidrule(lr){2-7}
& \multirow{2}{*}{0.5}
  & MAE  & -1.950 & 0.062 & 1 & 0.063 \\
& & RMSE & -2.677 & \textbf{0.028} & 1 & 0.063 \\
\cmidrule(lr){2-7}
& \multirow{2}{*}{0.7}
  & MAE  & -1.049 & 0.177 & 6 & 0.406 \\
& & RMSE & -0.422 & 0.347 & 8 & 0.594 \\
\midrule
\multirow{8}{*}{Gait}
& \multirow{2}{*}{0.1}
  & MAE  & -3.906 & \textbf{0.009} & 0 & \textbf{0.031} \\
& & RMSE & -4.500 & \textbf{0.005} & 0 & \textbf{0.031} \\
\cmidrule(lr){2-7}
& \multirow{2}{*}{0.3}
  & MAE  & -7.160 & \textbf{0.001} & 0 & \textbf{0.031} \\
& & RMSE & -1.814 & 0.072 & 3 & 0.156 \\
\cmidrule(lr){2-7}
& \multirow{2}{*}{0.5}
  & MAE  & -3.780 & \textbf{0.010} & 0 & \textbf{0.031} \\
& & RMSE & -3.177 & \textbf{0.017} & 0 & \textbf{0.031} \\
\cmidrule(lr){2-7}
& \multirow{2}{*}{0.7}
  & MAE  & -0.774 & 0.241 & 4 & 0.219 \\
& & RMSE & -1.396 & 0.118 & 2 & 0.094 \\
\bottomrule
\end{tabular}
\label{tab:significance}
\end{table}

\subsubsection{Prototype utilisation}

We analysed how many of the 32 prototypes were actually active at test time by computing the attention weight assigned to each of the 32 prototypes across the entire test sets’ windows. We report two summary statistics: (1) the number of prototypes individually receiving more than half of a uniform share of attention (i.e., average weight above 1/(2×32)); and (2) the number of prototypes needed to jointly account for 90\% of total attention mass. We present the results for missingness ratio 0.10 (seed 1) across all five datasets in Table~\ref{tab:active_prototypes} below.

We find that prototype utilisation is broad rather than concentrated on a small interpretable subset for each dataset. At least 31 of the 32 prototypes individually receive more than a half-share of attention, and between 27 and 29 of the 32 prototypes are needed to jointly account for 90\% of attention mass. This suggests the prototype bank is not over-provisioned at bank size 32. Stock and Gait are marginally more selective under the 90\%-cumulative measure (27 and 28 active prototypes respectively) than Weather, PhysioNet, and Beijing (29 each). 

\begin{table}[h!]
\centering
\caption{Number of active prototypes per dataset under the usage threshold and the 90\% cumulative usage criterion.}
\begin{tabular}{l|cc}
\toprule
Dataset & $n_{\text{active}}$ (threshold) & $n_{\text{active}}$ (90\% cumulative) \\
\midrule
Stock     & 31 & 27 \\
Weather   & 32 & 29 \\
PhysioNet & 32 & 29 \\
Beijing   & 32 & 29 \\
Gait      & 32 & 28 \\
\bottomrule
\end{tabular}
\label{tab:active_prototypes}
\end{table}

\subsection{Visualisations}
\begin{figure}[h]
\centering
\begin{subfigure}{0.95\linewidth}
    \centering
    \includegraphics[width=\linewidth]{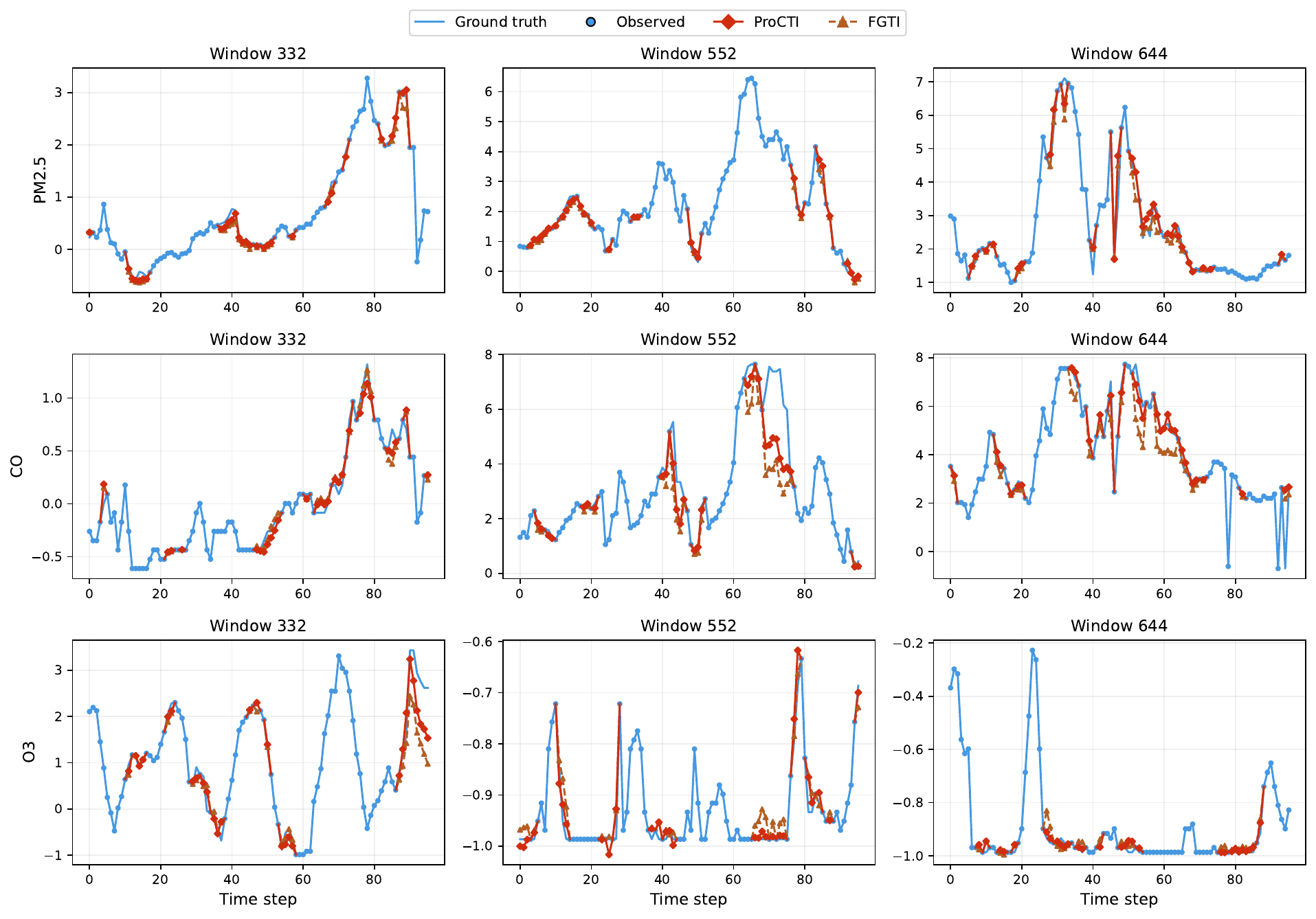}
    \caption{ProCTI vs FGTI}
\end{subfigure}

\vspace{0.5em}

\begin{subfigure}{0.95\linewidth}
    \centering
    \includegraphics[width=\linewidth]{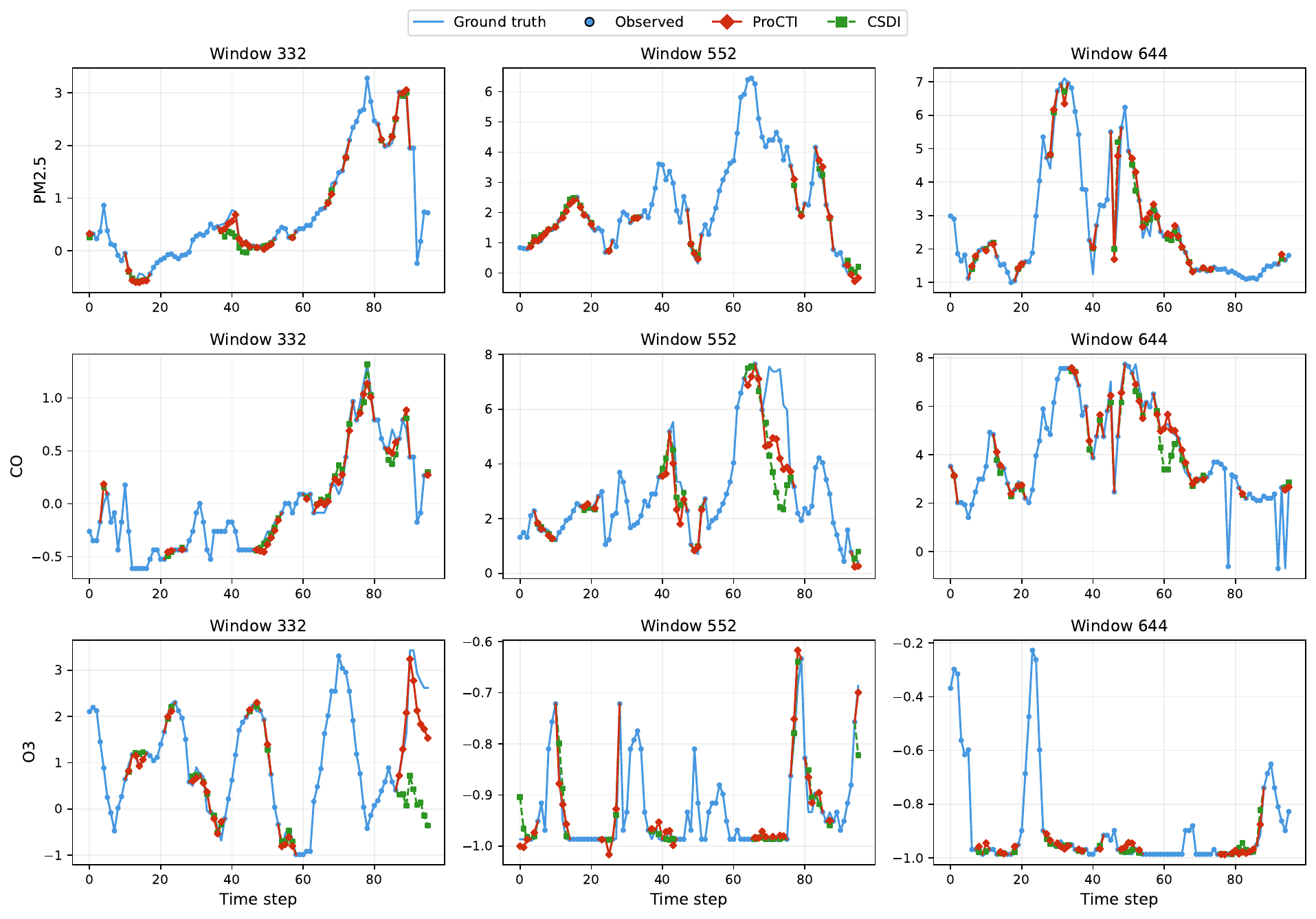}
    \caption{ProCTI vs CSDI}
\end{subfigure}

\caption{Representative imputation comparison on the Beijing dataset.}
\label{fig:beijing_imputation_examples}
\end{figure}

\begin{figure}[h]
\centering
\begin{subfigure}{0.95\linewidth}
    \centering
    \includegraphics[width=\linewidth]{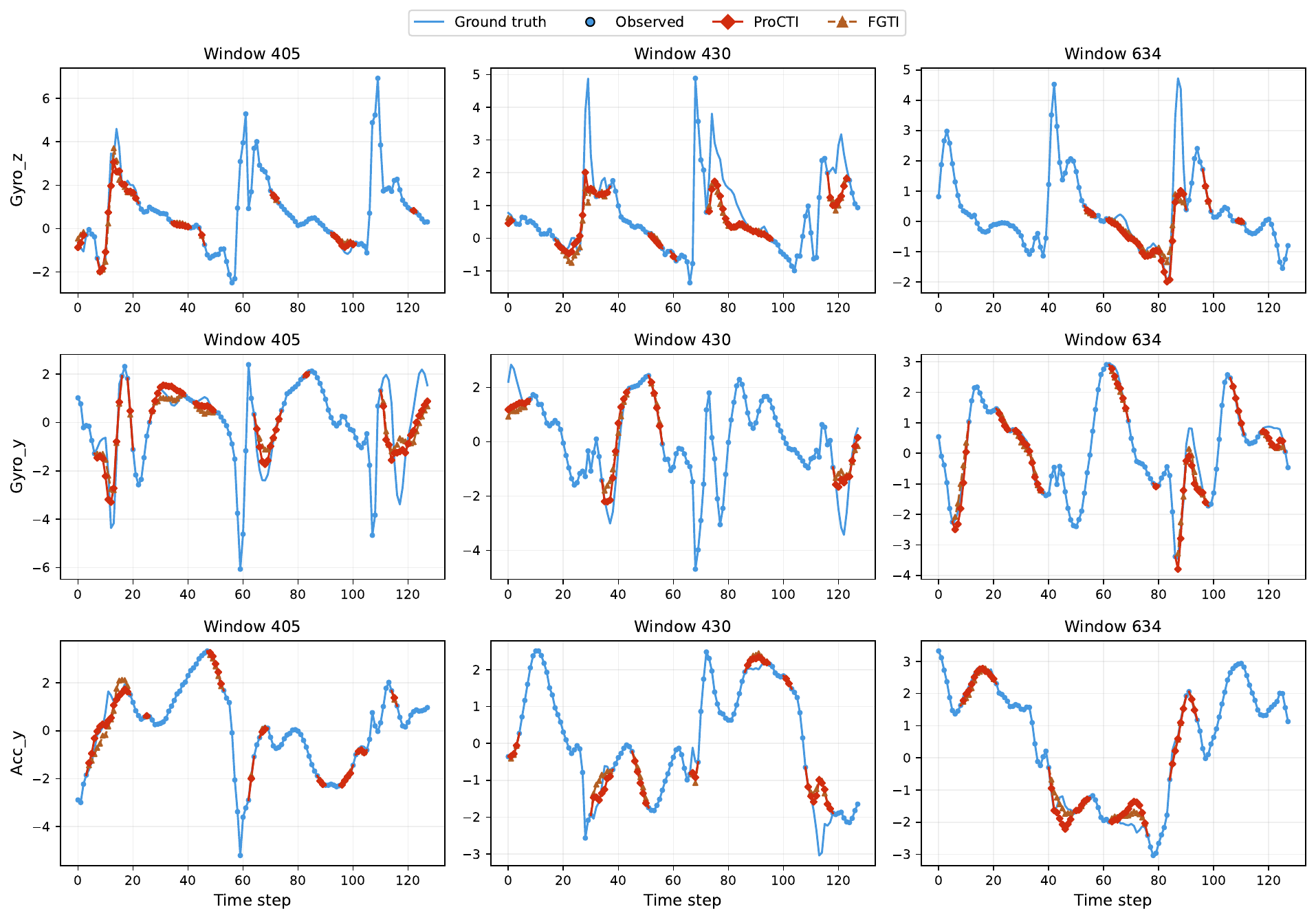}
    \caption{ProCTI vs FGTI}
\end{subfigure}

\vspace{0.5em}

\begin{subfigure}{0.95\linewidth}
    \centering
    \includegraphics[width=\linewidth]{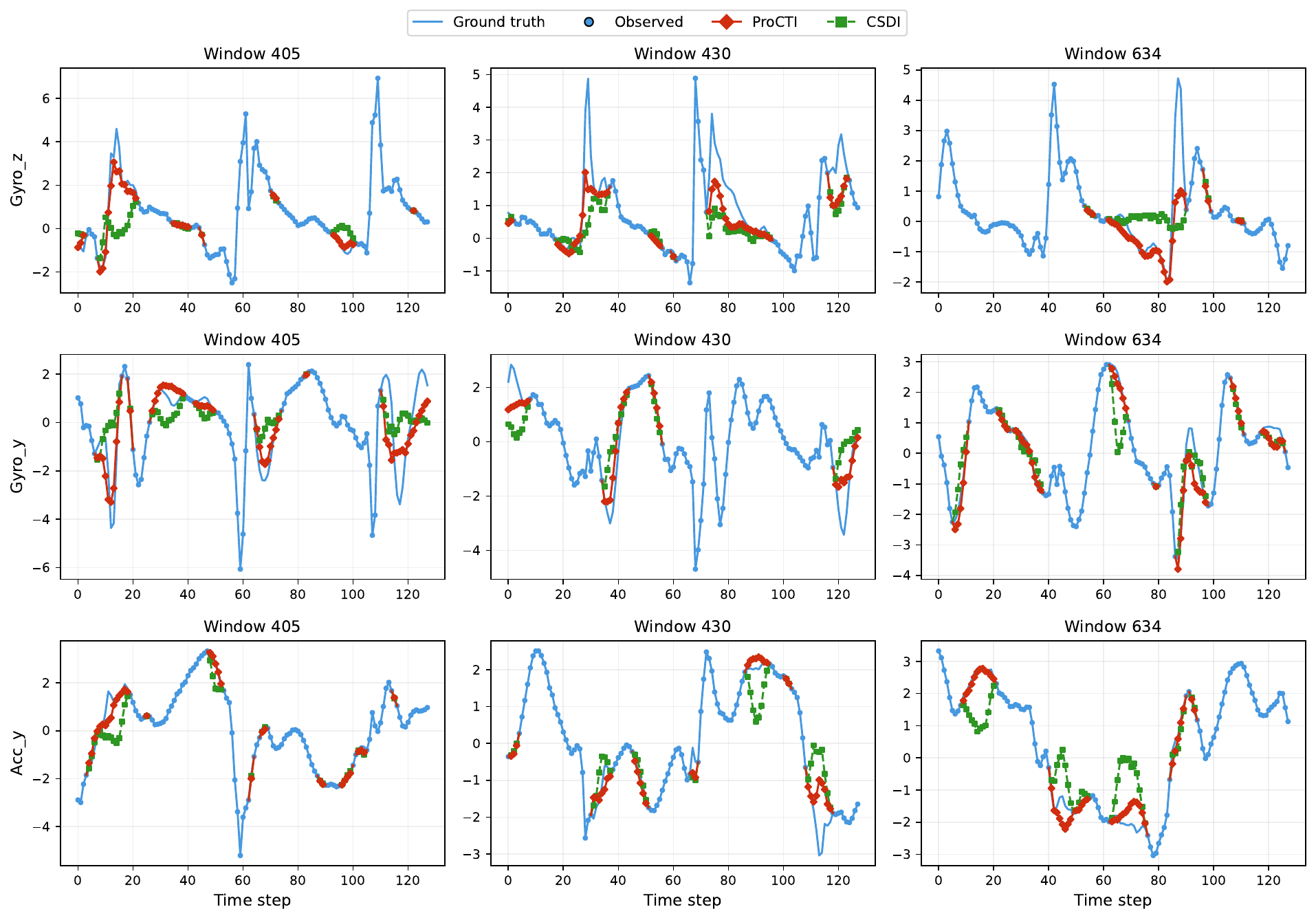}
    \caption{ProCTI vs CSDI}
\end{subfigure}

\caption{Representative imputation comparison on the Gait dataset.}
\label{fig:gait_imputation_examples}
\end{figure}

\begin{figure}[h]
\centering
\begin{subfigure}{0.95\linewidth}
    \centering
    \includegraphics[width=\linewidth]{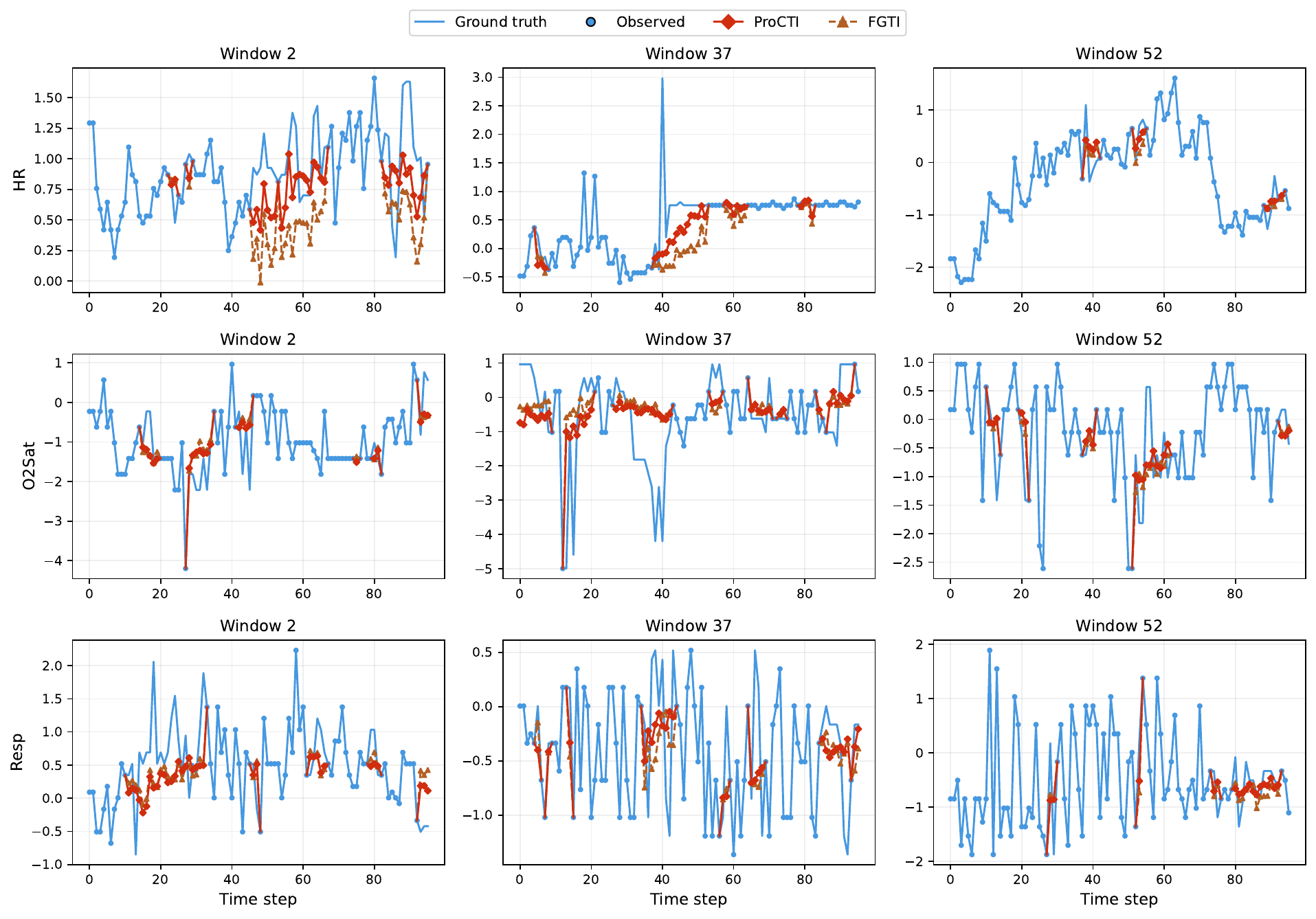}
    \caption{ProCTI vs FGTI}
\end{subfigure}

\vspace{0.5em}

\begin{subfigure}{0.95\linewidth}
    \centering
    \includegraphics[width=\linewidth]{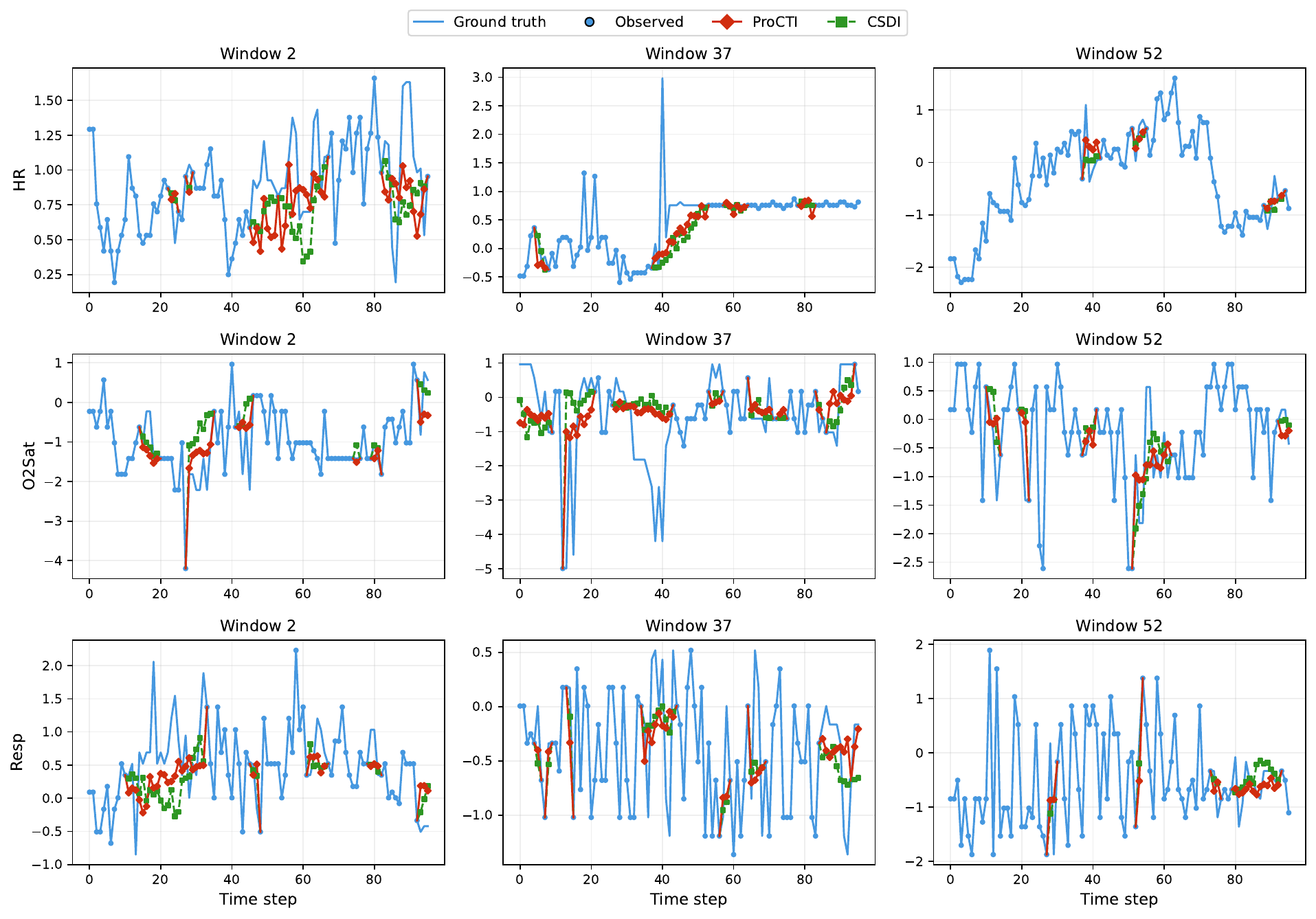}
    \caption{ProCTI vs CSDI}
\end{subfigure}

\caption{Representative imputation comparison on the PhysioNet dataset.}
\label{fig:physionet_imputation_examples}
\end{figure}

In Figure~\ref{fig:beijing_imputation_examples}, Figure~\ref{fig:gait_imputation_examples}, and Figure~\ref{fig:physionet_imputation_examples}, we present representative imputation examples for the Beijing, Gait, and PhysioNet datasets, comparing ProCTI, FGTI, and CSDI. Across all three datasets, CSDI imputations are more prone to deviating from the ground-truth trajectories, whereas ProCTI follows the true values within the masked regions more closely. ProCTI and FGTI often produce visually similar reconstructions, reflecting their shared frequency-aware diffusion conditioning framework. Nonetheless, closer inspection of several masked segments, together with the results in Table~\ref{tab:markovmask_results_aggregated} and Table~\ref{tab:markovmask_results}, shows that ProCTI consistently attains lower overall errors than FGTI, demonstrating the benefit of incorporating global conditioning signals. 

\begin{figure*}[t]
\centering

\begin{subfigure}{\textwidth}
    \includegraphics[width=\linewidth,page=1]{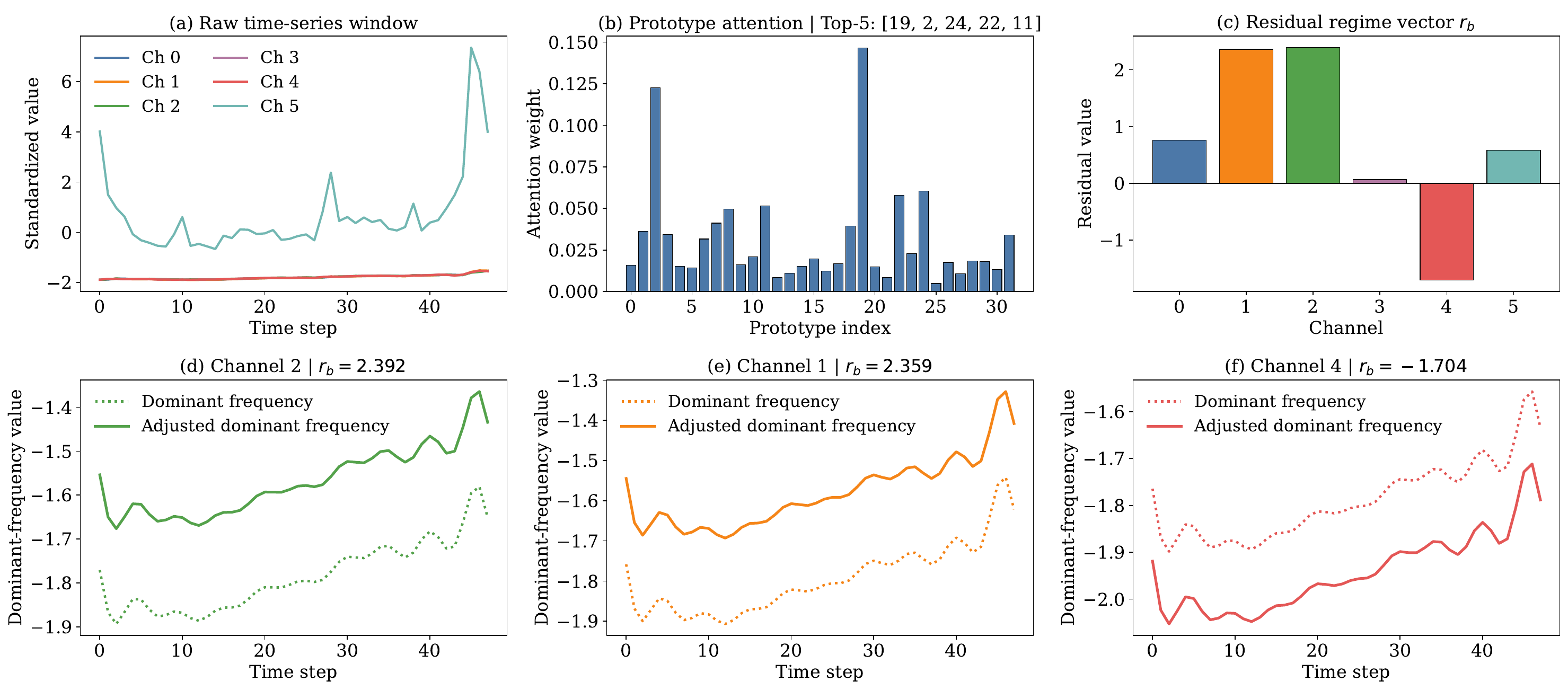}
    \caption{Window 0, Top-3 |$r_b$| channels: [2,1,4]}
\end{subfigure}

\begin{subfigure}{\textwidth}
    \includegraphics[width=\linewidth,page=1]{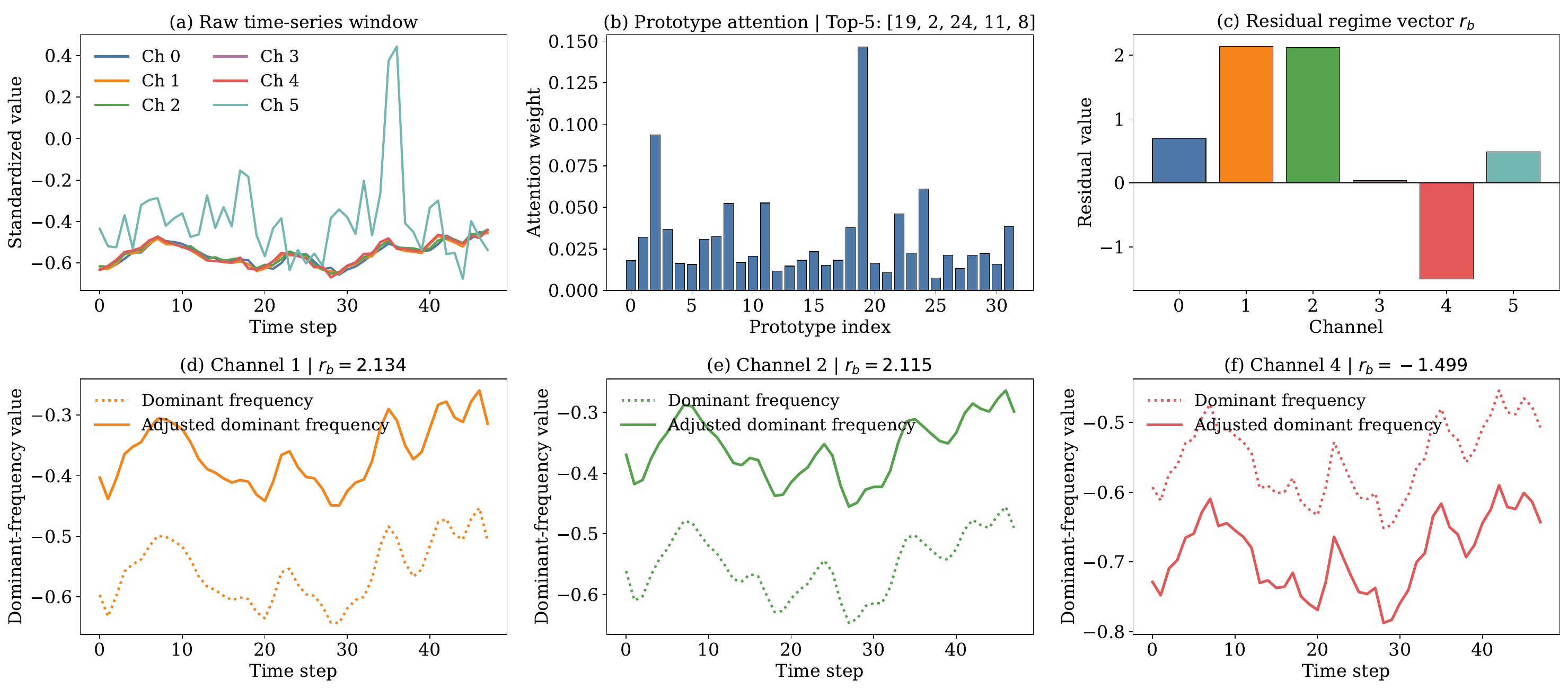}
    \caption{Window 25, Top-3 |$r_b$| channels: [1, 2, 4]}
\end{subfigure}

\begin{subfigure}{\textwidth}
    \includegraphics[width=\linewidth,page=1]{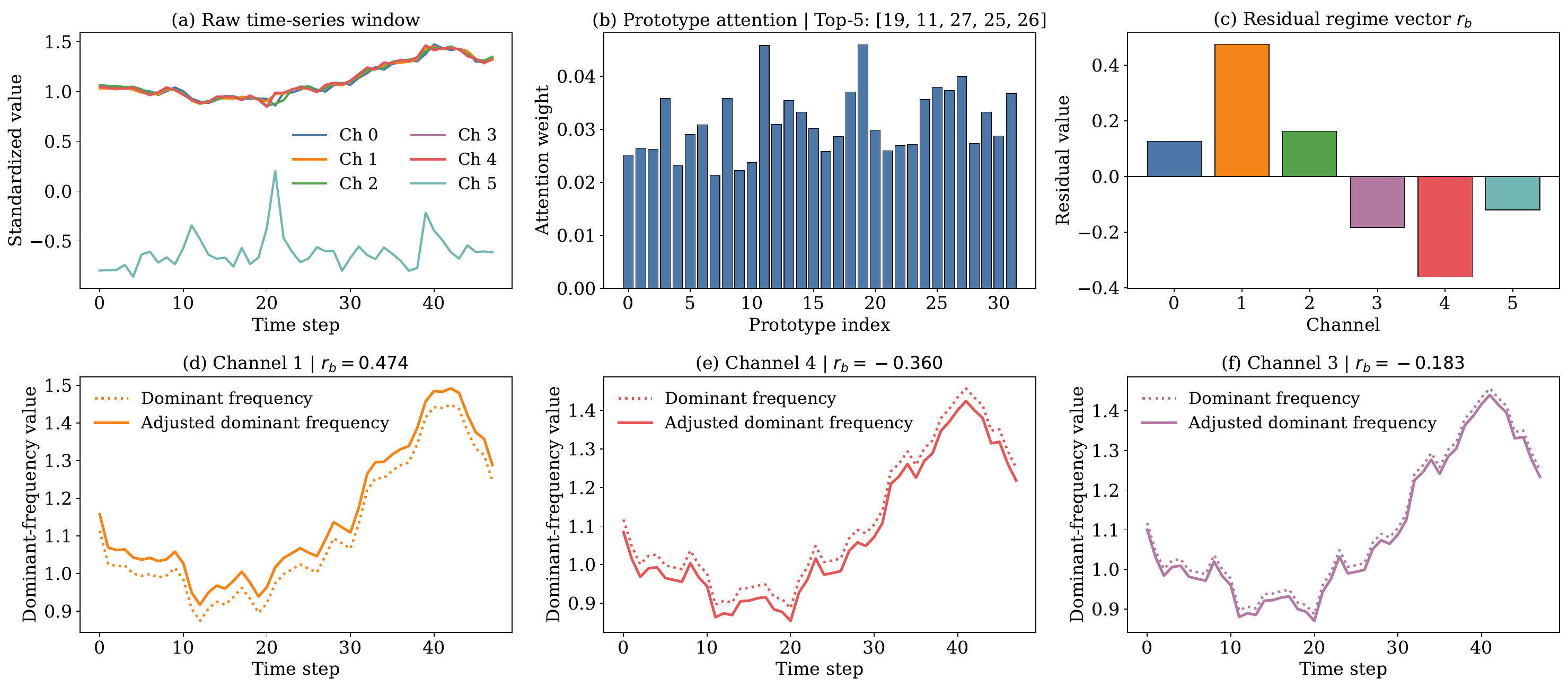}
    \caption{Window 45, Top-3 |$r_b$| channels: [1, 4, 3]}
\end{subfigure}

\caption{
Prototype usage visualisation examples on the Stock dataset for four representative windows. Each panel shows (a) the raw time-series window, (b) prototype attention weights, (c) the residual regime vector $r_b$, and (d)--(f) the dominant-frequency signal before and after prototype-based adjustment for the three channels with the largest $|r_b|$. The prototype attention weights and $r_b$ values are obtained after training for 50 epochs, at which point the learned scaling parameter is $\alpha = 0.0907$.
}
\label{fig:stock_window_examples}
\end{figure*}

Finally, Figure~\ref{fig:stock_window_examples} illustrates representative examples of prototype usage on window samples from the Stock dataset. Specifically, the figure shows the raw time series, the attention weights assigned by each window to the prototypes, the corresponding regime vector $r_b$ values, and the resulting adjustments to the dominant-frequency component of the three attributes with the largest absolute $r_b$ values.

\subsection{Broader impact statement}\label{appn:broaderimpacts}
The positive societal impacts of this work are directly related to time-series-data applications where missing or incomplete data are common, and improved imputation can positively affect downstream tasks. Example domains include healthcare, transportation, environmental monitoring, and industrial systems, where sensor readings or records may be missing or faulty. In healthcare, for instance, better recovery of missing clinical measurements may assist patient monitoring or diagnosis. On the other hand, potential unintended consequences in downstream applications due to incorrect imputations should also be considered. In such settings, technologies such as ProCTI should be used as decision-support tools rather than as replacements for human judgment, particularly in high-stakes scenarios.

